\documentclass{article}

\usepackage{microtype}
\usepackage{graphicx}
\usepackage{subcaption}

\usepackage{url}            % simple URL typesetting
\usepackage{amsfonts}       % blackboard math symbols
\usepackage{amsmath}
\usepackage{nicefrac}       % compact symbols for 1/2, etc.
\usepackage{microtype}      % microtypography
\usepackage{booktabs} % for professional tables

\usepackage{hyperref}

\usepackage[vlined,ruled]{algorithm2e}
\SetKwProg{Fn}{Function}{}{}
\SetKwFor{For}{for (}{)}{}
\SetKwInOut{Parameter}{parameter}

\usepackage[accepted]{icml2019}

\icmltitlerunning{Transfer Learning by Modeling a Distribution over Policies}

\begin{document}

\twocolumn[
\icmltitle{Transfer Learning by Modeling a Distribution over Policies}



\icmlsetsymbol{equal}{*}

\begin{icmlauthorlist}
\icmlauthor{Disha Shrivastava}{equal,udem}
\icmlauthor{Eeshan Gunesh Dhekane}{equal,udem}
\icmlauthor{Riashat Islam}{mcgill}
% \icmlauthor{Iaesut Saoeu}{ed}
% \icmlauthor{Fiuea Rrrr}{to}
% \icmlauthor{Tateu H.~Yasehe}{ed,to,goo}
% \icmlauthor{Aaoeu Iasoh}{goo}
% \icmlauthor{Buiui Eueu}{ed}
% \icmlauthor{Aeuia Zzzz}{ed}
% \icmlauthor{Bieea C.~Yyyy}{to,goo}
% \icmlauthor{Teoau Xxxx}{ed}
% \icmlauthor{Eee Pppp}{ed}
\end{icmlauthorlist}

\icmlaffiliation{udem}{Mila, Universit\'e de Montr\'eal}
\icmlaffiliation{mcgill}{Mila, McGill University}
% \icmlaffiliation{mila}{School of Computation, University of Edenborrow, Edenborrow, United Kingdom}

\icmlcorrespondingauthor{Disha Shrivastava}{dishu.905@gmail.com}
\icmlcorrespondingauthor{Eeshan Gunesh Dhekane}{eeshandhekane@gmail.com}

\icmlkeywords{Transfer Learning, Generative Modeling, Reinforcement Learning, ICML}

\vskip 0.3in
]



\printAffiliationsAndNotice{\icmlEqualContribution} % otherwise use the standard text.

\begin{abstract}
% Recent work named VFunc by~\cite{bachman2018vfunc} demonstrates the potential of addressing Bayesian deep learning in the function space rather than parameter space. This generative approach allows for learning with entropy-regularized objective maximization by forming a variational lower bound to the objective, resulting in learning of diverse policies that can be tweaked by latent variables. This feature motivates us to hypothesize that modeling distribution over policies, rather than the parameters of a single policy, can improve transfer of learning in different environments. In this project, we study the aspects in which VFunc improves transfer of learning across different environments. We conduct experiments on fully-observable GridWorld ~\footnote{Easy MDP: https://github.com/zafarali/emdp} and partially observable MiniGrid~\cite{gym_minigrid} environments. In each setting, we compare with vanilla policy-gradient approaches like REINFORCE and A2C and demonstrate that training with VFunc on the source environment always results in better transfer of learning in the target environment in terms of faster convergence to the maximum reward. 

Exploration and adaptation to new tasks in a transfer learning setup is a central challenge in reinforcement learning. In this work, we build on the idea of modeling a distribution over policies in a Bayesian deep reinforcement learning setup to propose a transfer strategy. 
% We hypothesize that maximizing the entropy over a distribution of policies can be useful in transfer learning. 
Recent works have shown to induce diversity in the learned policies by maximizing the entropy of a distribution of policies~\cite{bachman2018vfunc,garnelo2018neural} and thus, we postulate that our proposed approach leads to faster exploration resulting in improved transfer learning. We support our hypothesis by demonstrating favorable experimental results on a variety of settings on fully-observable GridWorld\footnote{Easy MDP: https://github.com/zafarali/emdp} and partially observable MiniGrid~\cite{gym_minigrid} environments.

% Recent work named VFunc\,\cite{bachman2018vfunc} demonstrates the potential of addressing Bayesian deep learning in the function space rather than parameter space. This generative approach allows for learning with entropy-regularized objective maximization by forming a variational lower bound to the objective, resulting in learning of diverse policies. We hypothesize that the approach of exploration in the function space leads to better encoding of policies, which can significantly improve transfer of learning. In this paper, we demonstrate and experimentally establish the aspects in which VFunc improves transfer of learning across different environments. We conduct experiments on fully-observable GridWorld\!~\footnote{Easy MDP: https://github.com/zafarali/emdp} and partially observable MiniGrid~\cite{gym_minigrid} environments. In each setting, we compare VFunc with vanilla policy-gradient approaches like REINFORCE and A2C and demonstrate that training with VFunc on the source environment always results in better transfer of learning in the target environment in terms of faster convergence to the maximum reward.
\end{abstract}

% \vspace*{-0.55cm}
\section{Introduction}
% Motivation for doing transfer with VFunc. Why do we think it is better? Importance of variational approach and entropy term

% Bayesian deep learning has been a useful idea to capture predictive uncertainty. Bayesian deep neural networks are useful for representing uncertainty over model classification \cite{}, active learning \cite{} and for exploration in RL \cite{}. Recent work has further shown that instead of capturing uncertainty over parameters, the uncertainty over functions can more useful \cite{}. 

% In this work, we present a Bayesian deep learning framework, building on existing work \cite{} to demonstrate transfer learning in RL. We show that maximizing entropy over a distribution over functions can be useful for transfer. 

%  

Reinforcement learning (RL) has had major success recently, in terms of mastering highly complex games like Go~\cite{silver2017mastering} and robotic control tasks. However, the ability of agents to transfer across tasks still remains an important problem to be solved towards artificial general intelligence. Transfer learning plays 
% an important
a crucial role in this; to give the agents the ability to adapt across tasks. In this work, we postulate that a key step towards transfer and multi-task learning is for the agent to be able to adapt and explore faster in the new task. Our goal is for the agent to explore unseen regions of the state space quickly in new tasks by re-using its past experiences. However, doing this in the on-policy setting is difficult, since the agent cannot re-use its past data as the task changes. Towards this, we provide an approach for faster exploration in the new task. 
%The ability to transfer relies on how quickly the agent can adapt and explore in the new task, so as to familiarize itself towards the goal seeking challenge in the new task. 

We present a transfer learning strategy which fundamentally relies on Bayesian deep learning and the ability to represent a distribution over functions, as in \cite{bachman2018vfunc} \cite{garnelo2018neural}. Bayesian methods rely on modeling the uncertainty over value functions to represent the agent's belief
% and uncertainty over
of the environment. Recent work has shown that neural networks can be used to represent an uncertainty over the space of all possible functions \cite{bachman2018vfunc}.
% , and this framework can be used in the reinforcement learning setup. 
% By modeling a distribution over functions, 
The idea of modeling a distribution over functions can be adapted in the RL setting to model a distribution over policies, such that we can also maximize the entropy over this distribution of policies. This is similar to maximum entropy exploration in RL, where instead of local entropy maximization, recent work 
%has been shown to 
maximizes the global entropy over the space of all possible sub-optimal policies. 

% This idea naturally induces a distribution over policies, where each sampled policy from the distribution can lead to highly diverse policies and diverse trajectories. We develop our approach based on this intuition to lead to faster exploration.

Our work relies on modeling a distribution over policies for transfer learning, where the pre-trained policies can be useful for learning the target policy in the transfer setup.
% to maximize coverage of the state space. 
We hypothesize that if for a given task, we can maximize the entropy of the distribution over policies, then this model can be useful for transfer. This is because, in the new tasks, each sampled policy would lead to a highly diverse trajectory, leading to faster coverage of the state space.

We build upon a recent work named VFunc~\cite{bachman2018vfunc}, which presents a framework to represent a distribution over functions rather than the space of parameters. VFunc models the distribution $p(f, z)$, where $f$ is an element of the function space and $z$ is the latent variable. VFunc not only mitigates the intractable modeling of uncertainty in the parameter space but also leads to an efficient method of sampling of functions.
%from the function space with the marginalization of the latent variable. 
\begin{figure*}[!t]
    \centering
    \includegraphics[width = \textwidth]{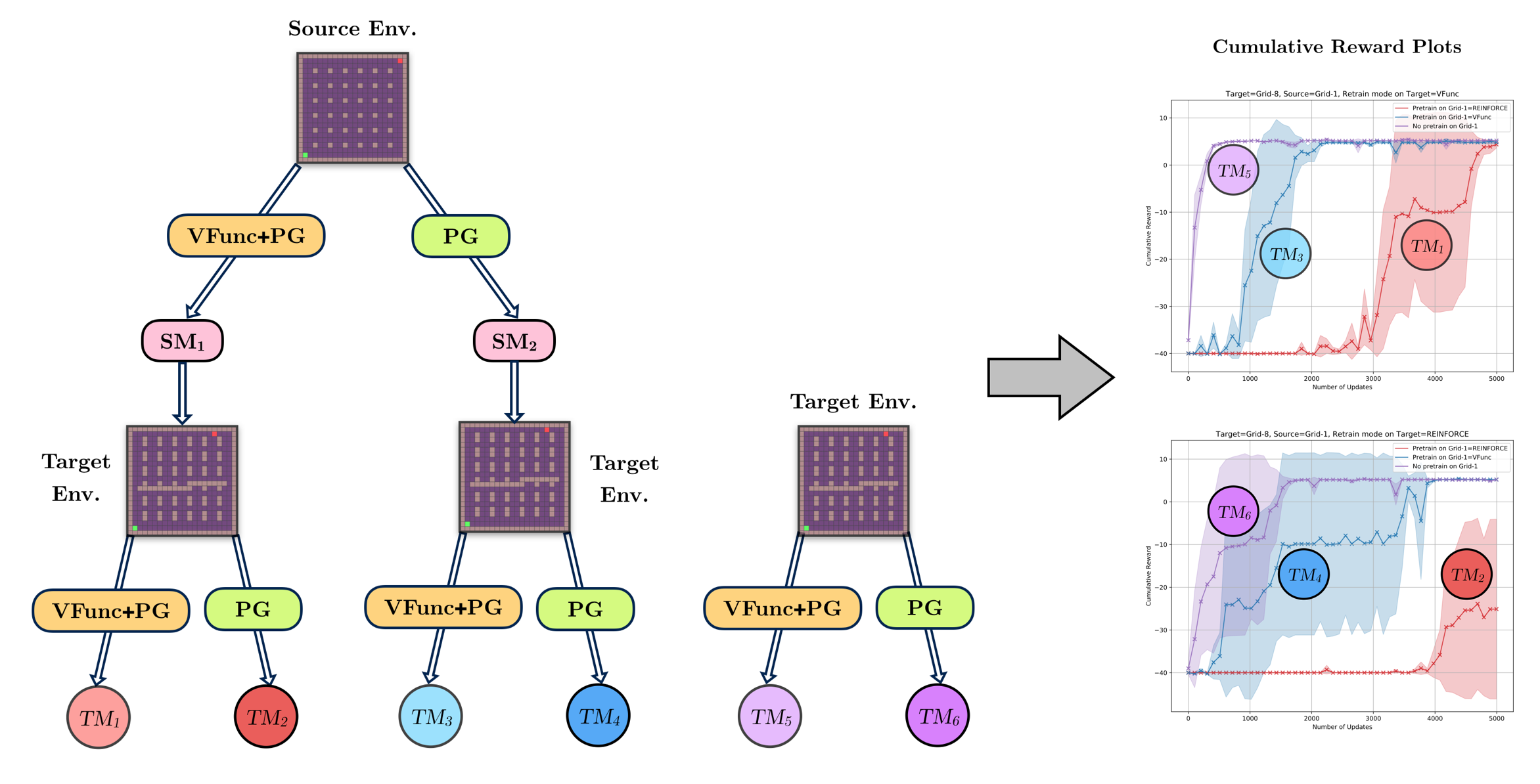}
    \caption{\textbf{Schematic explaining Transfer Learning with VFunc:} We start with training the source environment using both VFunc+Policy Gradient (PG) and just Policy Gradient (PG) algorithms. PG algorithms correspond to either REINFORCE or A2C. The Source Model (SM) is then fine-tuned in the target environments to produce the Transferred Model (TM). We also train the target environment from scratch. Performance in terms of cumulative reward plots is then observed for all TM's. }
    \label{fig:formulation:arch}
\end{figure*}
% Note that many learning problems in the areas of Bayesian deep learning and RL can be cast as
VFunc can be viewed as the maximization of the entropy-regularized objectives of the form: $\mathcal{L}(f) = \mathbb{E}_{f\sim p(f)}\left[R(f)\right] + \lambda \cdot H(f)$. Here, $R(f)$ is a measure of the ``goodness'' of the function $f$, and $H(f)$ represents the entropy of $f\sim p(f)$. $R(f)$ can be the data log-likelihood in case of Bayesian deep learning or the reward in case of RL. Note that the efficient sampling of functions $f\sim p(f)$, by marginalizing $z$ from $p(f, z)$ in VFunc, can be directly used towards maximization of these objectives. Further, the joint distribution $p(f, z)$ can be utilized to obtain a variational lower bound for the entropy term $H(f)$. Thus, the approach of modeling $p(f, z)$ can be applied in an efficient and simple manner to a large number of Bayesian deep learning and RL problems.
%via variational lower bound maximization of entropy-regularized objectives. 
In reinforcement learning, 
%the prediction function
% $p(f)$ is 
%chosen to be
% the distribution over policies. 
% Thus, note that the above approach can be seen as entropy-regularized reward maximization in the policy space. 
the aim is to maximize a reward term $\mathbb{E}_{\pi\sim p(\pi)}\left[R(\pi)\right]$, measuring the goodness of the policy $\pi$,
subject to large exploration in the policy space ensured by the term $H(p(\pi))$, which represents the entropy of the distribution over policies. Note that by varying the latent variable $z$, we can obtain different policies.

To summarize, we propose a transfer learning approach based on modeling a distribution over policies.
%aspects in which the approach of  improves transfer of learning for different environments.
The entropy term in the optimization objective forces the model to consider a diverse set of functions while training. 
% This training with exploration results in the learning of better encoding of policies in terms of the distribution over policy. This distribution can be carried over to new problem settings, giving a better pre-training prior, which can then be fine-tuned for the new setting. Thus, we are motivated to hypothesize that modeling the distribution over policies can improve the transfer of learning in different environment settings. 
Since the learned policy parameters in the train (source) task are representative of the policy distribution, it improves the exploration of trajectories in the test (target) task. This offers benefits as compared to learning the target policy based on random explorations as the knowledge gained from the source task can be efficiently transferred to the target task to improve performance.
% Instead of relying on random exploration of trajectories, which would be the case if VFunc-based pre-trained parameters are not incorporated, the learning of policy involves better and faster trajectory exploration owing to the knowledge gained from the train task.
Extensive experimentation on GridWorld
% \!\footnote{Easy MDP: https://github.com/zafarali/emdp} 
and MiniGrid demonstrates that transfer to target environments by training with VFunc on the source environment not only results in improved convergence but also leads to diversity in learned policies.

\section{Related Work}
Significant work has been done investigating various approaches for transfer learning~\cite{taylor2009transfer,lazaric2012transfer}. These include meta-learning based transferable strategies~\cite{gupta2018meta}, learning decision states with information bottleneck for transfer~\cite{goyal2019infobot}, attention-based deep architectures for selective transfer~\cite{rajendran2015attend}, semi-Markov Decision Processes based approach~\cite{mehta2008transfer}, distillation based approach~\cite{distral}, and successor features based transfer~\cite{barreto2017successor}. 
In contrast, our approach utilizes diverse exploration in policy space via variational lower bound maximization of entropy-regularized objectives as the basis of transfer. 
There has also been some work that considers the problem of exploration strategies~\cite{schmidhuber1991curious,osband2016deep,houthooft2016variational}.
However, these approaches do not carry out transfer of learning from source task to target tasks. 
% We demonstrate that our approach leads to successful transfer of prior knowledge. 

In addition, recent works~\cite{haarnoja2018soft,neu2017unified,nachum2017bridging} demonstrate the advantages of entropy-regularized approaches. Also, there has been significant work on modeling policies with latent variables~\cite{hausman2018learning,florensa2017stochastic,haarnoja2018latent,eysenbach2018diversity}. In particular, the contribution of~\cite{hausman2018learning} is the policy gradient formulation for hierarchical policies with task-conditional latent variables. 
On the other hand, we rely on optimization of the entropy-regularized objectives for inducing diversity with latent variable based modeling of distribution over policies.  

\section{Method Description}
In this section, we provide a brief description of VFunc \cite{bachman2018vfunc}, a deep generative model for modeling a distribution over the function space. Later, we describe how we use VFunc to do transfer learning.

\subsection{Modeling Distribution over Policies}
VFunc \cite{bachman2018vfunc} consists of a prediction network for a stochastic mapping of inputs to outputs and a recognition network for encoding a function into a latent variable. The prediction network $p(y\mid x, z)$ maps the inputs $x$ to outputs $y$ conditioned on latent variable $z$. In the case of RL problems, the underlying function $f$ is the 
%REVIEW: distribution over the policy space
policy, the inputs $x$ are the state representations and the outputs $y$ are the corresponding actions. The space of the latent variable $z$ is assigned a latent variable prior, denoted by $p(z)$. Analogously, the space of the functions (policies) $f$ is assigned a function prior $\bar{p}(f)$. The recognition network $q(z\mid f)$ learns a latent variable encoding for policy $f$. Further, the variational lower bound on the entropy term is formulated using the recognition network. The optimization objective which allows estimation of the true posterior $p(f\mid D)$ given dataset $D$, is given as follows (taken directly from \cite{bachman2018vfunc}):
\begin{equation}
    \label{eq:formulation:optObjective}
    \textbf{Max} \, \, \mathcal{L}(f) = \mathbb{E}_{f\sim p(f)}[
    R(D\mid f) + \log \bar{p}(f)] + \lambda H(f)
\end{equation}
The data-dependent reward term $\mathbb{E}_{f\sim p(f)}[R(D\mid f)]$ can be computed via standard policy-gradient techniques like A2C or REINFORCE. The function prior $\bar{p}(f)$ can be chosen to be unnormalized energy, i.e., arbitrary regularizer function. The estimation of $H(f)$ is not so straight-forward. However, it can be achieved by variational lower bound given below (taken directly from \cite{bachman2018vfunc}):
\begin{equation}
    \label{eq:formulation:varLowBound}
    H(f) \geq H(z) + \mathbb{E}_{f, z\sim p(f, z)}[\log q(z\mid f)] + H(f\mid z)
\end{equation}
The latent variable prior $p(z)$ is chosen to have simple estimation of $H(z)$. \cite{bachman2018vfunc} model $p(y\mid x, z)$ to be a Gaussian distribution so that $H(f\mid z)$ estimation becomes easier. Finally, the conditional cross-entropy term can be computed in terms of recognition network $q(z\mid f)$ and the underlying joint distribution $p(f, z)$. Thus, from using the lower bound from Eqn.\ref{eq:formulation:varLowBound}, we can compute a lower bound estimate of the optimization objective.

VFunc models a function $f$ as a set of $K$ input-output pairs, called as a partially observed version of the function, denoted by $\hat{f}$. To generate a (partially observed) policy function given a latent variable, they sample $K$ state representations $\{x_k\}_{k = 1}^K$ from the space of inputs and sample actions $y_k \sim p(y\mid x_k, z)$. Thus, the partially observed policy $\hat{f}$ is given by the input-output pairs: $\hat{f} = \{(x_k, y_k)\}_{k = 1}^K$. On the other hand, to encode a partially observed policy $\hat{f} = \{(x_k, y_k)\}_{k = 1}^K$, they use the prediction network and a fixed default latent variable $\bar{z}$ to obtain the default output $\hat{y}_k$. Then, the loss function for each input-output pair is computed as $L(y_k, p(y\mid x_k, \bar{z}))$. Back-propagating the sum of all such loss functions gives a differential gradient $\sum_{k = 1}^K\nabla_{\bar{z}}L(y_k, p(y\mid x_k, \bar{z}))$, which is an indicative of how skewed the prediction network is when the fixed default latent variable is used for prediction. This gradient forms the input to a multi-layer feed-forward network that essentially forms $q(z\mid \hat{f})$ and predicts a latent variable $z$ for $\hat{f}$.

\begin{algorithm}%[H]
\footnotesize
\DontPrintSemicolon
\SetAlgoLined
\SetKwInOut{Input}{Input}
\SetKwInOut{Output}{Output}
\SetKwInOut{Nothing}{}
% \Input{\quad Training algorithm $\mathcal{A}\in\{\text{VFunc}, \text{REINFORCE}\}$}
\Input{\quad Training algorithm $\mathcal{A}\in\{\text{PG}, \text{VFunc+PG}\}$}
% \Nothing{\quad PG denotes REINFROCE, A2C.}
\Input{\quad Initial Policy $\pi_\theta(a\mid s, z)$, parameterized with $\theta$}
\Input{\quad Train and Test task distributions: $P_{\mathrm{train}}, P_{\mathrm{test}}$}
\Output{\quad Trained Policy $\pi_{\theta^{\dagger}}(a\mid s, z)$ for the test task}
\BlankLine 
\SetAlgoLined
\Fn{Transfer ($\mathcal{A}, \pi_\theta(a\mid s, z), P_{\mathrm{train}}, P_{\mathrm{test}}$)}{

    \vspace*{7pt}
    $\bullet$\ \textbf{Training Setup:}
    \vspace*{3pt}

        $\theta\leftarrow$\ random initialization
    
        Sample train task $\tau_{\mathrm{train}}\sim P_{\mathrm{train}}$
    
        \For {\text{episodes $ = 1:N_{\mathrm{train}}$}}{
            
            Train policy $\pi_{\theta}$ on task $\tau_{\mathrm{train}}$ with algorithm $\mathcal{A}$
            
        }
        \textbf{end for}
    
    $\theta^\ast\leftarrow$\ Parameters after training on $\tau_{\mathrm{train}}$
% For loop working. Add ";" at end is not needed but a blank line, it should work fine!
%\BlankLine

    \vspace*{7pt}
    $\bullet$\ \textbf{Transfer Setup:}
    \vspace*{3pt}

    $\theta \leftarrow \theta^\ast$\ (transfer) \textbf{or} $\theta\leftarrow$ random init. (train from scratch)

%\BlankLine
    
%     \For {\text{episodes $ = 1:N_{target}$}}{
%         Sample a configuration $C$ based on $P_{source}(T)$${}^{\dagger}$
        
%         Sample a task $T \sim P_{target}(T)$ conditioned on C 
    
%         Update the policy parameters $\phi$ using Eqn. \ref{eq:formulation:optObjective} and training with $\mathcal{V}_{\mathcal{A}}\left({\lambda_{target}}\right)$ to get $\pi_\phi^{*}$
% }
%     \textbf{end for}
    
    Sample test task $\tau_{\mathrm{test}}\sim P_{\mathrm{test}}$
    
    \For {\text{episodes $ = 1:N_{\mathrm{test}}$}}{
            
            Train policy $\pi_{\theta}$ on task $\tau_{\mathrm{test}}$ with algorithm $\mathcal{A}$
            
    }
    \textbf{end for}
    
    $\theta^\dagger\leftarrow$\ Parameters after training on $\tau_{\mathrm{test}}$

}

\BlankLine
\Return $\pi_{\theta^{\dagger}}(a\mid s, z)$\;
\BlankLine

% \textbf{Remark:}${}^{\dagger}$ $C$ = position of goal and walls for \textit{GridWorld};\\
% $C$ = \#rooms, size of room, position of start and goal for \textit{MiniGrid}
\caption{Transfer Learning Via VFunc}
\label{algorithm}
\end{algorithm}

\subsection{Transfer Learning with VFunc}
For transfer learning with VFunc, we consider a set of tasks and the train, test task distributions\footnote{Here, train and test task distributions refer to source and target task distributions, respectively.}: $P_{\mathrm{train}}, P_{\mathrm{test}}$. In the training phase, a task $\tau_{\mathrm{train}}$ is sampled from $P_{\mathrm{train}}$. The parametrized policy $\pi_{\theta}$ is trained on the train task $\tau_{\mathrm{train}}$ with the VFunc algorithm~\cite{bachman2018vfunc} (using Eq.~\ref{eq:formulation:optObjective}) to obtain trained parameters $\theta^\ast$. Note that the training promotes exploration in the policy space due to the entropy term $H(f)$ in the objective~\ref{eq:formulation:optObjective}. Thus, the learned parameters encode the knowledge of the policy distribution. 

For transfer, we sample a new test task $\tau_{\mathrm{test}}$ from $P_{\mathrm{test}}$. For training a policy on the test task, we initialize the policy parameters with $\theta^\ast$. Since the pre-trained parameters represent a good and diverse encoding of policy distribution, it improves the exploration of trajectories in the test task. Instead of relying on random exploration of trajectories, which would be the case if VFunc based pre-trained parameters are not incorporated, the learning of policy involves better and faster trajectory exploration owing to the knowledge gained from the train task. Thus, we expect to observe faster learning and faster convergence to the maximum rewards in the test task. The results of the experiments, as described in the subsequent section, indeed validate this hypothesis. 
% To establish the better transfer due to VFunc, we also train another policy on the train task without VFunc. We use the learned parameters in this case for transfer to the test task.
Our transfer learning algorithm is described in Algorithm~\ref{algorithm} and its schematic representation is depicted in Figure~\ref{fig:formulation:arch}.

\section{Implementation}
In order to test our hypothesis, we carried out multiple experiments on different environments and variable settings within each environment. In these experiments, we compared training with VFunc to training with vanilla policy gradient-approaches. We provide the details of the environments used and various experimental settings below.

\subsection{Environments Used}
\subsubsection{GridWorld}
\begin{figure}
    \centering
    \begin{subfigure}[b]{0.15\textwidth}
        \includegraphics[width=\textwidth]{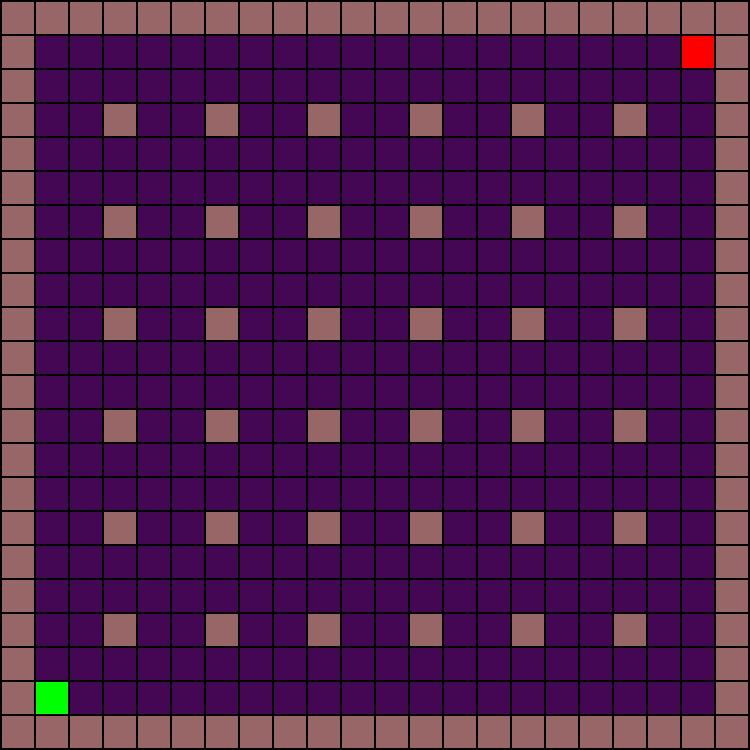}
        \caption{Grid-1}
        \label{fig:grid-1}
    \end{subfigure}
    ~ %add desired spacing between images, e. g. ~, \quad, \qquad, \hfill etc. 
    \begin{subfigure}[b]{0.15\textwidth}
        \includegraphics[width=\textwidth]{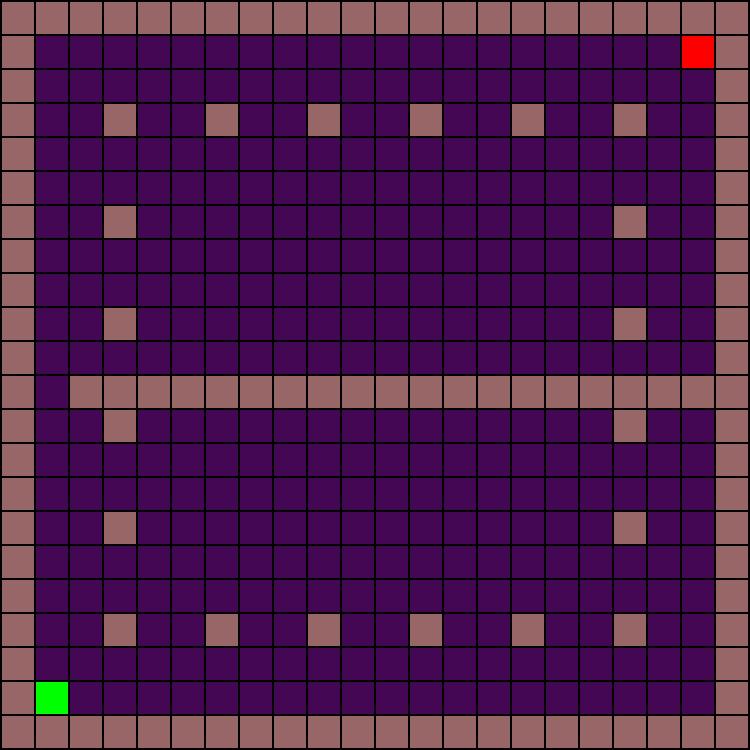}
        \caption{Grid-2}
        \label{fig:grid-2}
    \end{subfigure}
    ~ %add desired spacing between images, e. g. ~, \quad, \qquad, \hfill etc. 
    \begin{subfigure}[b]{0.15\textwidth}
        \includegraphics[width=\textwidth]{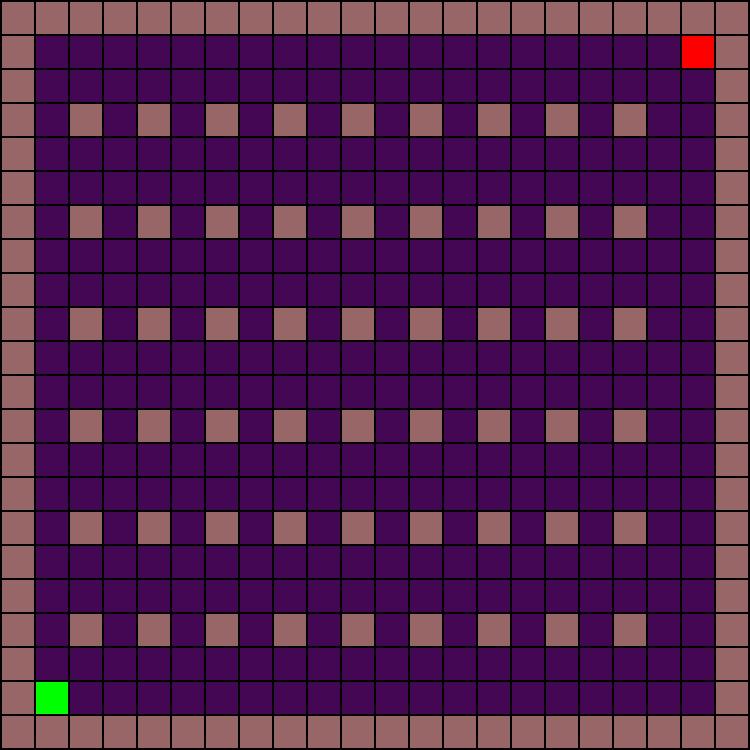}
        \caption{Grid-3}
        \label{fig:grid-3}
    \end{subfigure}
    ~
    \begin{subfigure}[b]{0.15\textwidth}
        \includegraphics[width=\textwidth]{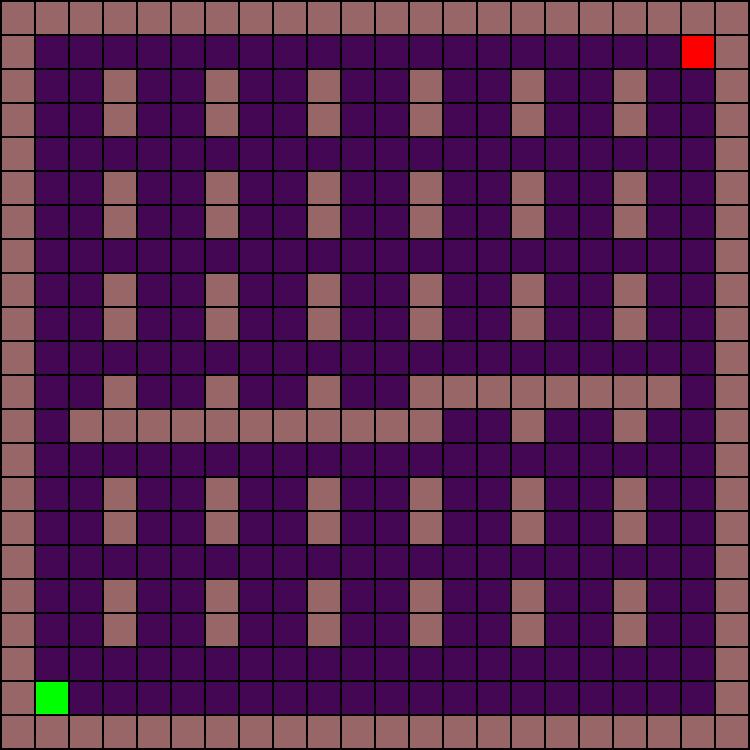}
        \caption{Grid-6}
        \label{fig:grid-6}
    \end{subfigure}
    ~
    \begin{subfigure}[b]{0.15\textwidth}
        \includegraphics[width=\textwidth]{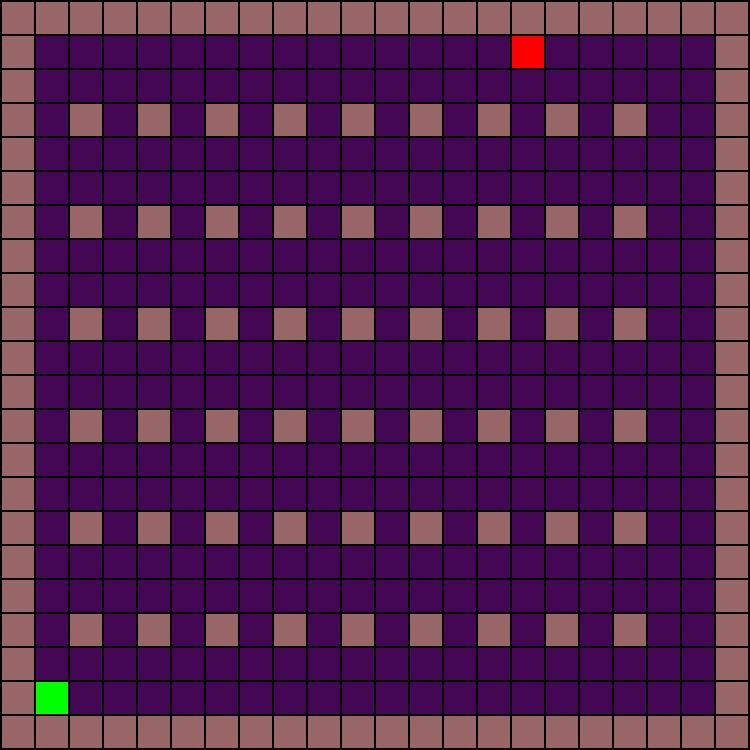}
        \caption{Grid-7}
        \label{fig:grid-7}
    \end{subfigure}
    ~
    \begin{subfigure}[b]{0.15\textwidth}
        \includegraphics[width=\textwidth]{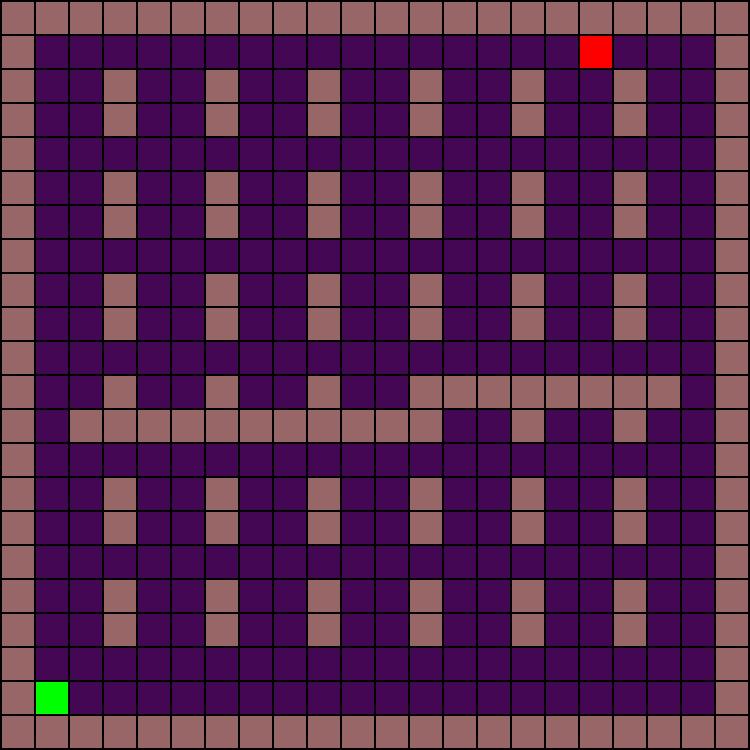}
        \caption{Grid-8}
        \label{fig:grid-8}
    \end{subfigure}
    \caption{Different GridWorld Environments Used. Red and Green squares correspond to start and goal positions, respectively. The Grid-1 is used as the \textbf{Source} environment and the rest as the \textbf{Target} environments.}\label{fig:gridworld}
\end{figure}

In this class of fully-observable environments, the task is to reach from a start position to a goal position in a rectangular grid. There is a reward of $+1$ on reaching the goal and a small penalty for the number of steps taken to reach the goal. Actions are deterministic and correspond to movement in the four directions. We experimented with six square gridworld domains of size 20 implemented with EasyMDP\footnote{Easy MDP: https://github.com/zafarali/emdp}. The details of each are provided in Figure \ref{fig:gridworld}.

\subsubsection{MiniGrid}
We also experimented with the partially-observable class of gridworld gym environments called MiniGrid~\cite{gym_minigrid}. Specifically, we used the Multi-Room Environment suite with 2 and 3 rooms of sizes 4 and 6. This environment has a series of connected rooms with doors that must be opened in order to get to the next room. The final room has the green goal square the agent must reach in order to get a reward of +1. There is a small penalty added for the number of steps to the reach the goal. There are two settings which we experimented with. In the first setting [\textbf{\textit{Dynamic}}], a different environment with the same configuration is chosen randomly at the beginning of each episode. The number of rooms and size of each room remains constant, but the positions of rooms, the way they are connected and the position of doors is dynamic. These factors make this suite of environments extremely difficult to solve using RL alone. In the second setting [\textbf{\textit{Static}}], the environment at the beginning of each episode is kept same by fixing the seed.

%%%%%%%%%%%%%%%%%%%%%%%%%%%%%%%%%%%%%%%%%%%%%%%%%%
%%%%%%%%%%%%%%%%%%%%%%%%%%%%%%%%%%%%%%%%%%%%%%%%%%
\begin{figure*}[h]
    \centering
    % ROW 1
    \begin{subfigure}[h]{0.495\textwidth}
        \begin{subfigure}[b]{0.495\textwidth}
            \includegraphics[trim = {1.0cm 0 1.5cm 0}, clip, width=\textwidth]{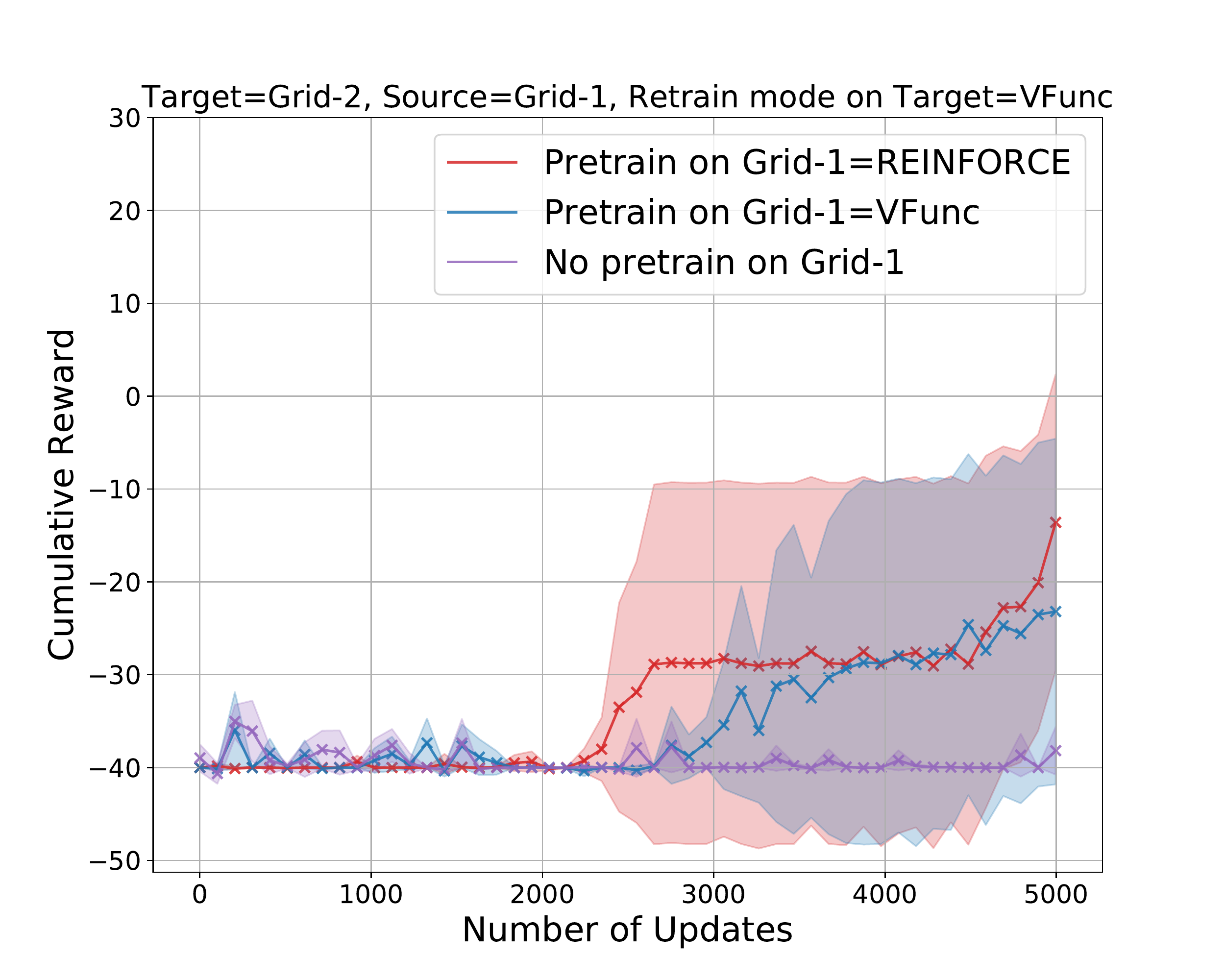}
            % \caption{Retrain with VFunc}
            \label{fig:vfunc_grid_2}
        \end{subfigure}
        \begin{subfigure}[b]{0.495\textwidth}
            \includegraphics[trim = {1.0cm 0 1.5cm 0}, clip, width=\textwidth]{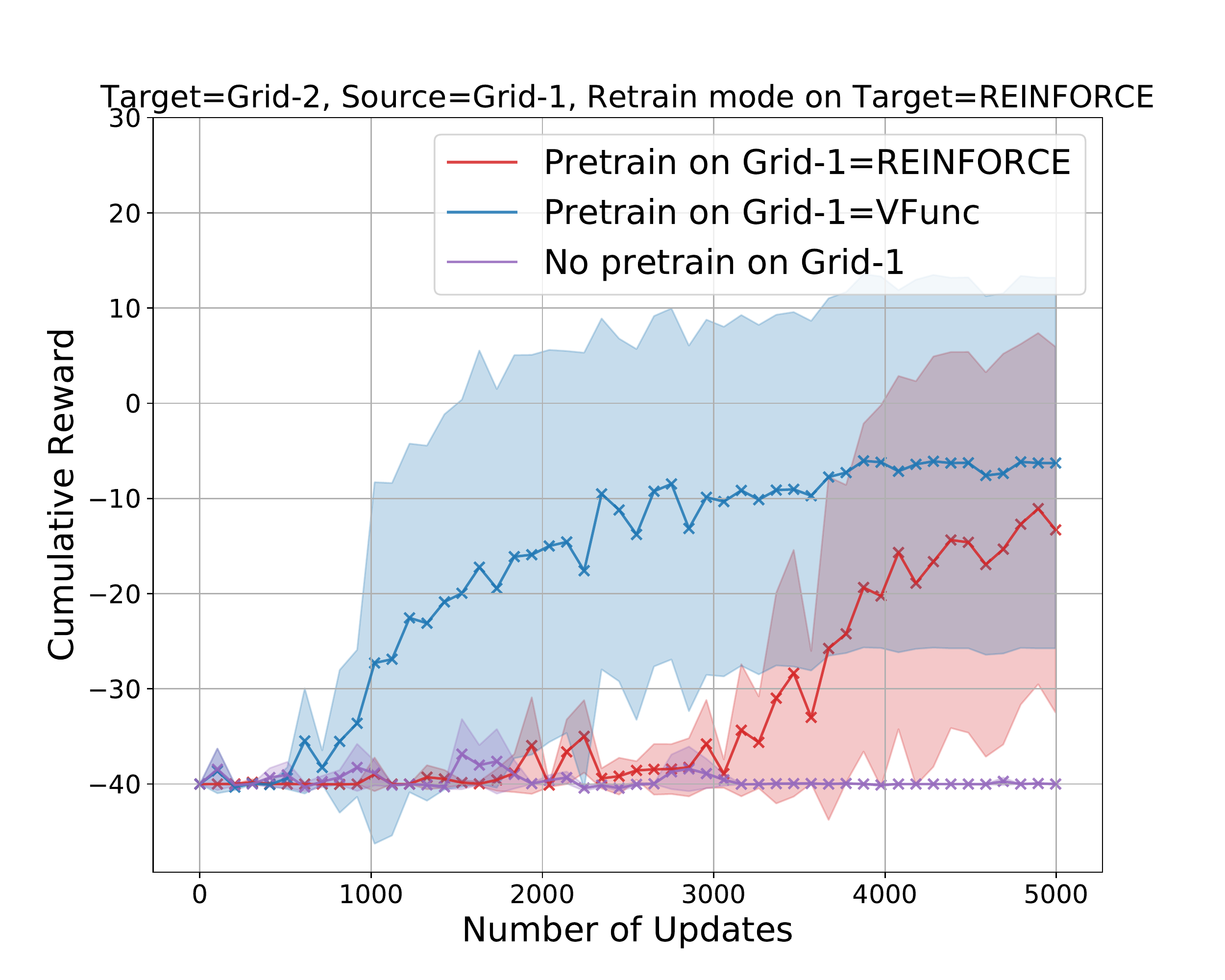}
            % \caption{Retrain with REINFORCE}
            \label{fig:r_grid_2}
        \end{subfigure}
        \vspace*{-1cm}
        \caption{Transfer on Grid-2}
        \label{transfer_grid_2}
    \end{subfigure}
    \centering
    \begin{subfigure}[h]{0.495\textwidth}
        \begin{subfigure}[b]{0.495\textwidth}
            \includegraphics[trim = {1.0cm 0 1.5cm 0}, clip, width=\textwidth]{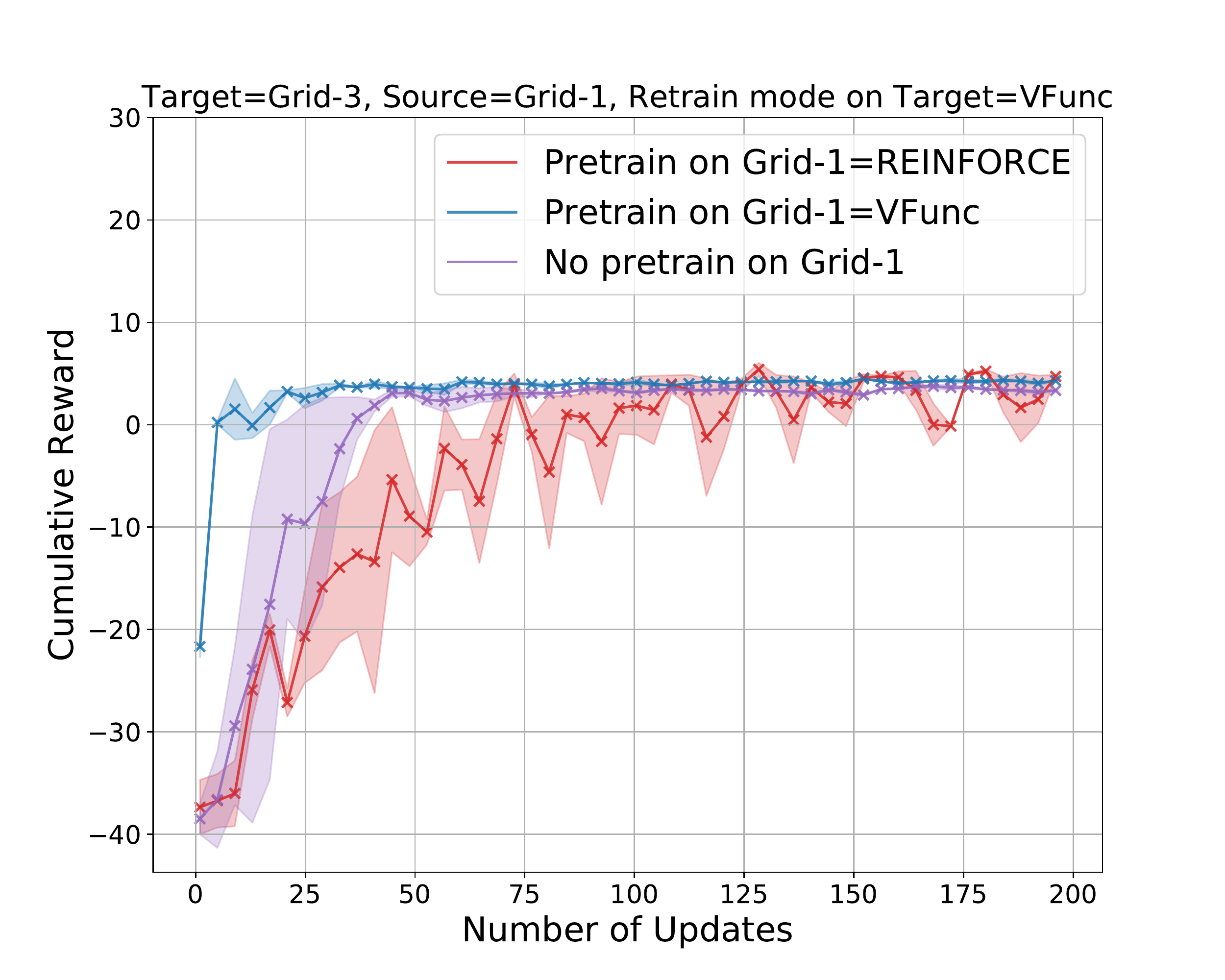}
            % \caption{Retrain with VFunc}
            \label{fig:vfunc_grid_3}
        \end{subfigure}
        \begin{subfigure}[b]{0.495\textwidth}
            \includegraphics[trim = {1.0cm 0 1.5cm 0}, clip, width=\textwidth]{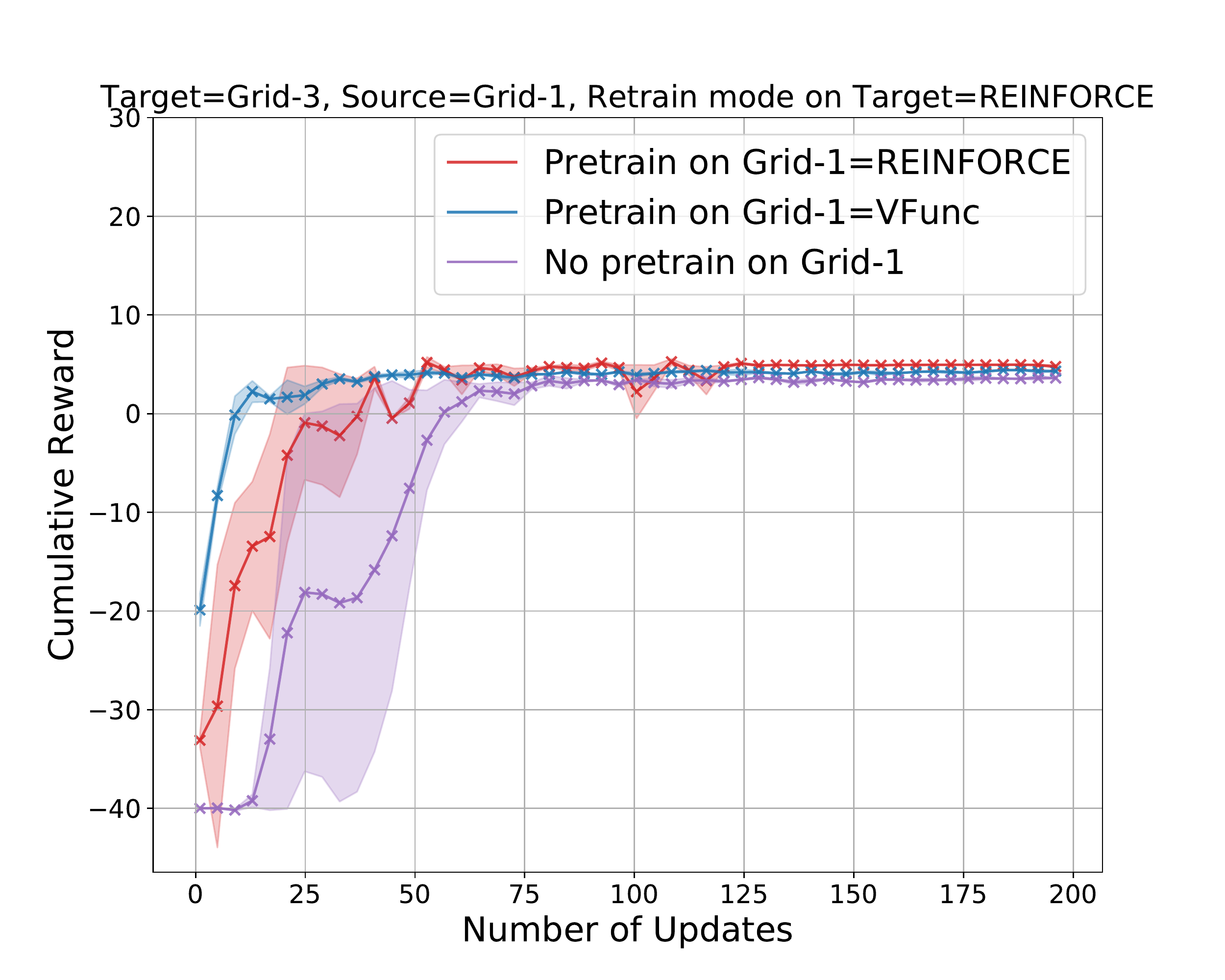}
            % \caption{Retrain with REINFORCE}
            \label{fig:r_grid_3}
        \end{subfigure}
        \vspace*{-1cm}
        \caption{Transfer on Grid-3}
         \label{transfer_grid_3}
    \end{subfigure}
    % ROW 2
    \bigskip
    \centering
     \begin{subfigure}[h]{0.495\textwidth}
        \begin{subfigure}[b]{0.495\textwidth}
            \includegraphics[trim = {1.0cm 0 1.5cm 0}, clip, width=\textwidth]{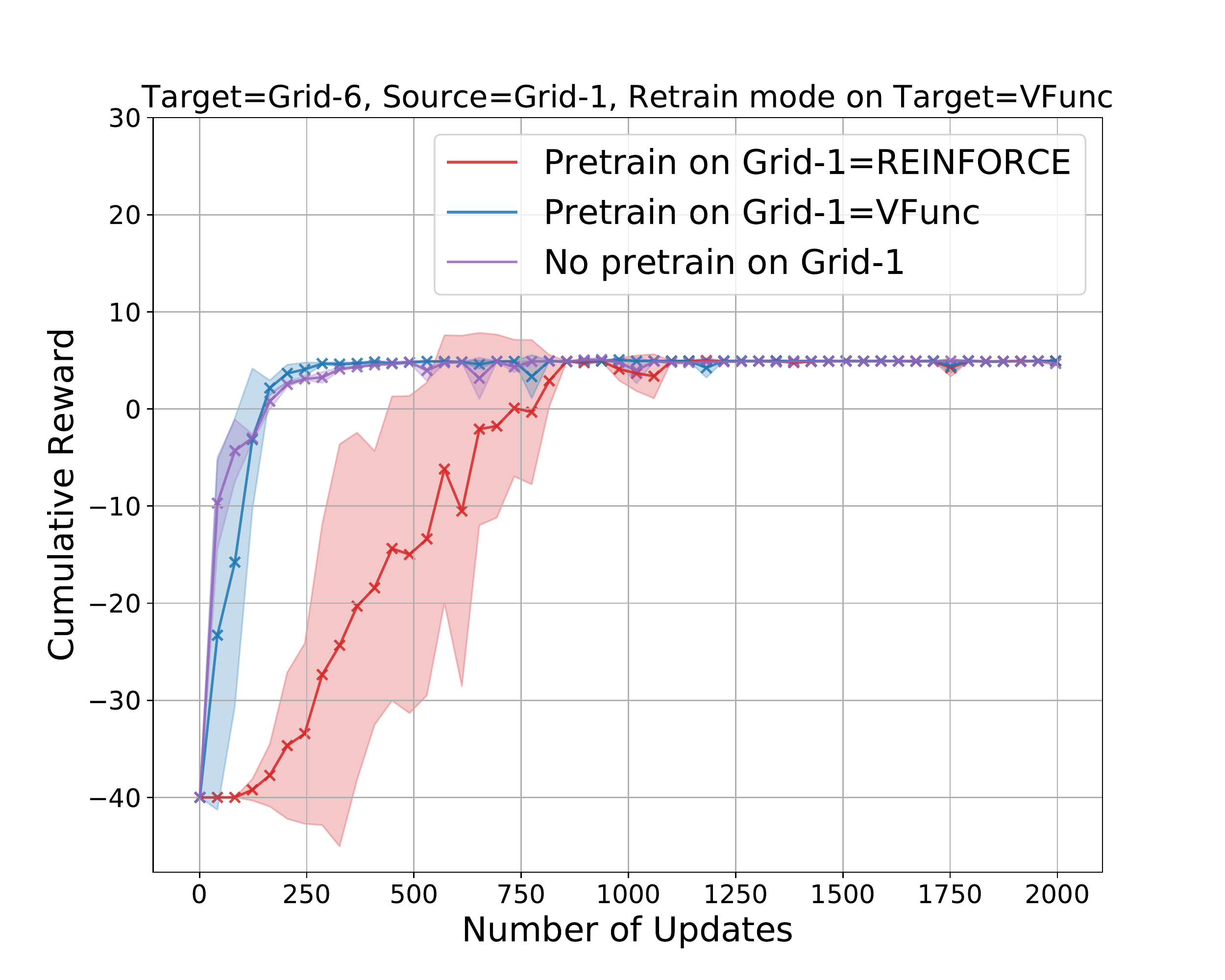}
            % \caption{Retrain with VFunc}
            \label{fig:vfunc_grid_3}
        \end{subfigure}
        \begin{subfigure}[b]{0.495\textwidth}
            \includegraphics[trim = {1.0cm 0 1.5cm 0}, clip, width=\textwidth]{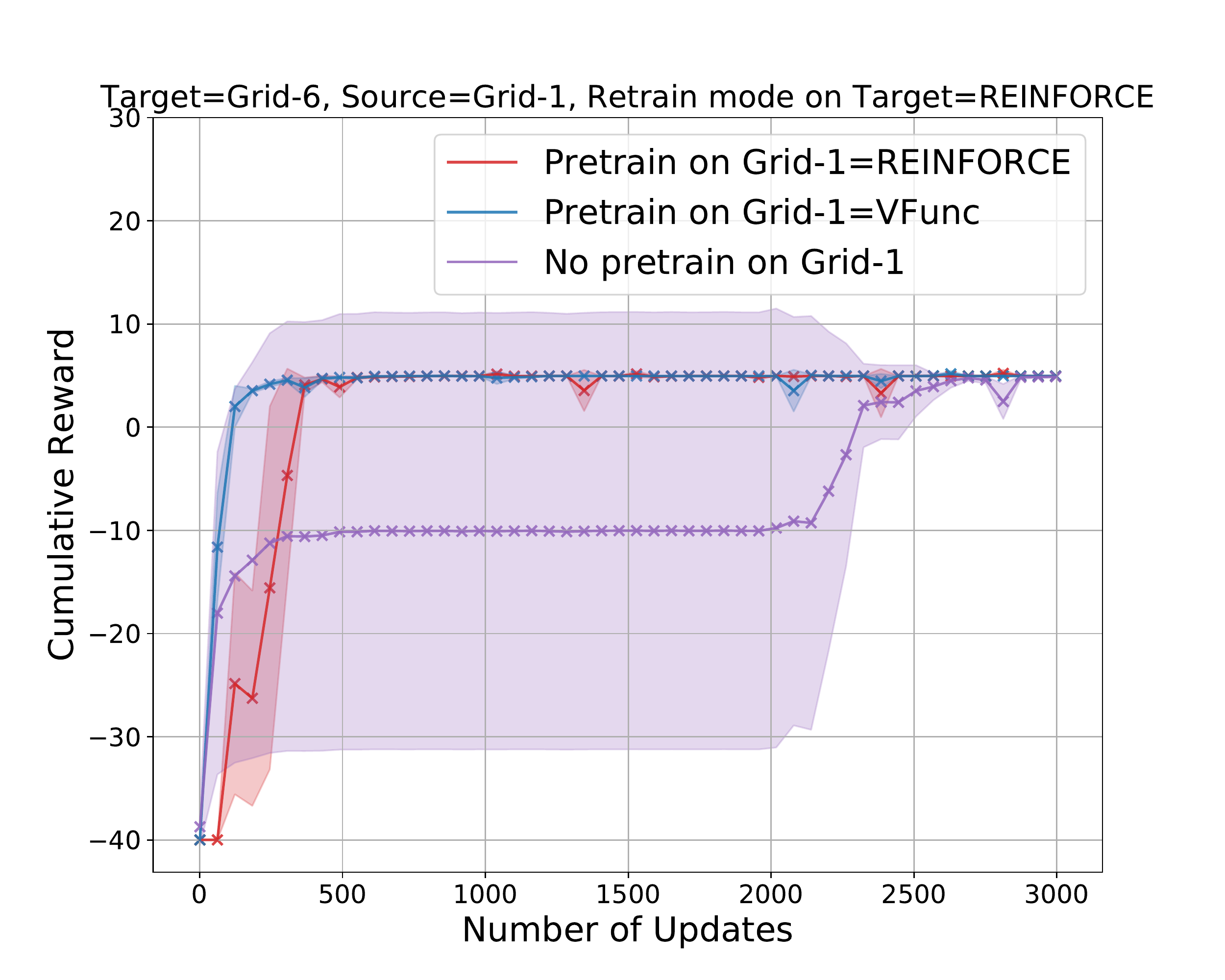}
            % \caption{Retrain with REINFORCE}
            \label{fig:r_grid_3}
        \end{subfigure}
        \vspace*{-1cm}
        \caption{Transfer on Grid-6}
         \label{transfer_grid_6}
    \end{subfigure}
    \centering
    \begin{subfigure}[h]{0.495\textwidth}
        \begin{subfigure}[b]{0.495\textwidth}
            \includegraphics[trim = {1.0cm 0 1.5cm 0}, clip, width=\textwidth]{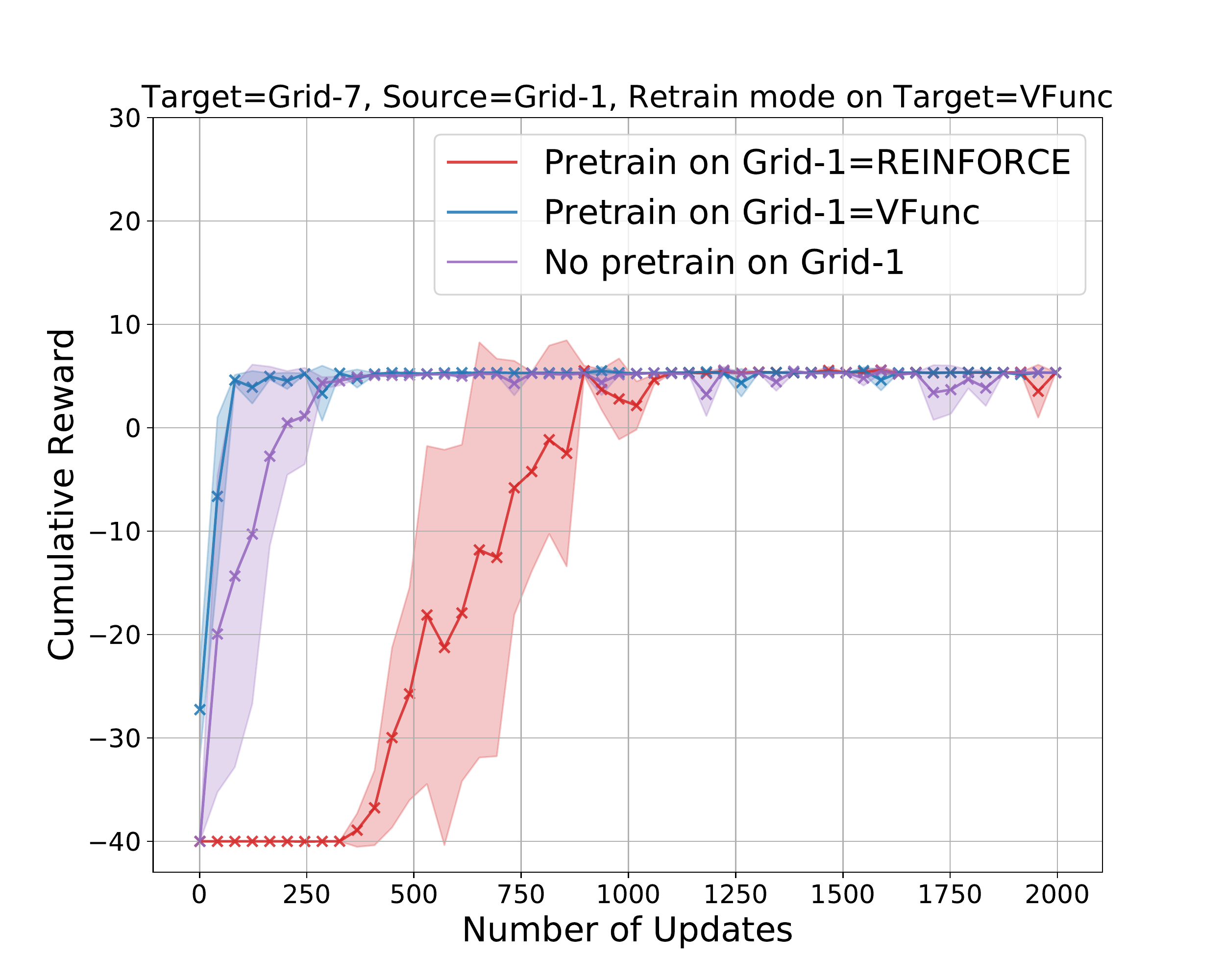}
            % \caption{Retrain with VFunc}
            \label{fig:vfunc_grid_3}
        \end{subfigure}
        \begin{subfigure}[b]{0.495\textwidth}
            \includegraphics[trim = {1.0cm 0 1.5cm 0}, clip, width=\textwidth]{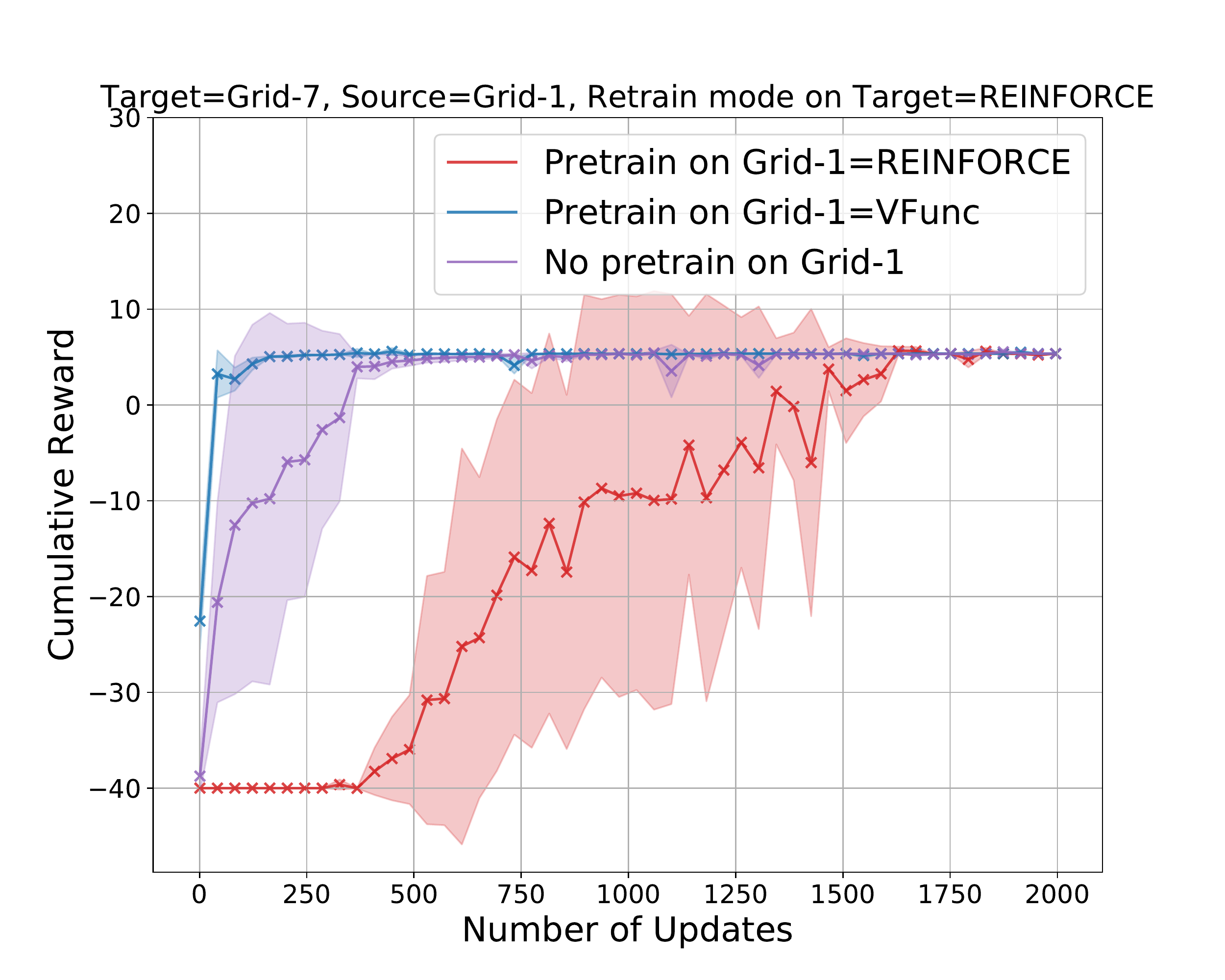}
            % \caption{Retrain with REINFORCE}
            \label{fig:r_grid_3}
        \end{subfigure}
        \vspace*{-1cm}
        \caption{Transfer on Grid-7}
         \label{transfer_grid_7}
    \end{subfigure}
    % ROW 3
    \bigskip
    \centering
    \vspace*{-0.85cm}
    \begin{subfigure}[h]{0.495\textwidth}
        \begin{subfigure}[b]{0.495\textwidth}
            \includegraphics[trim = {1.0cm 0 1.5cm 0}, clip, width=\textwidth]{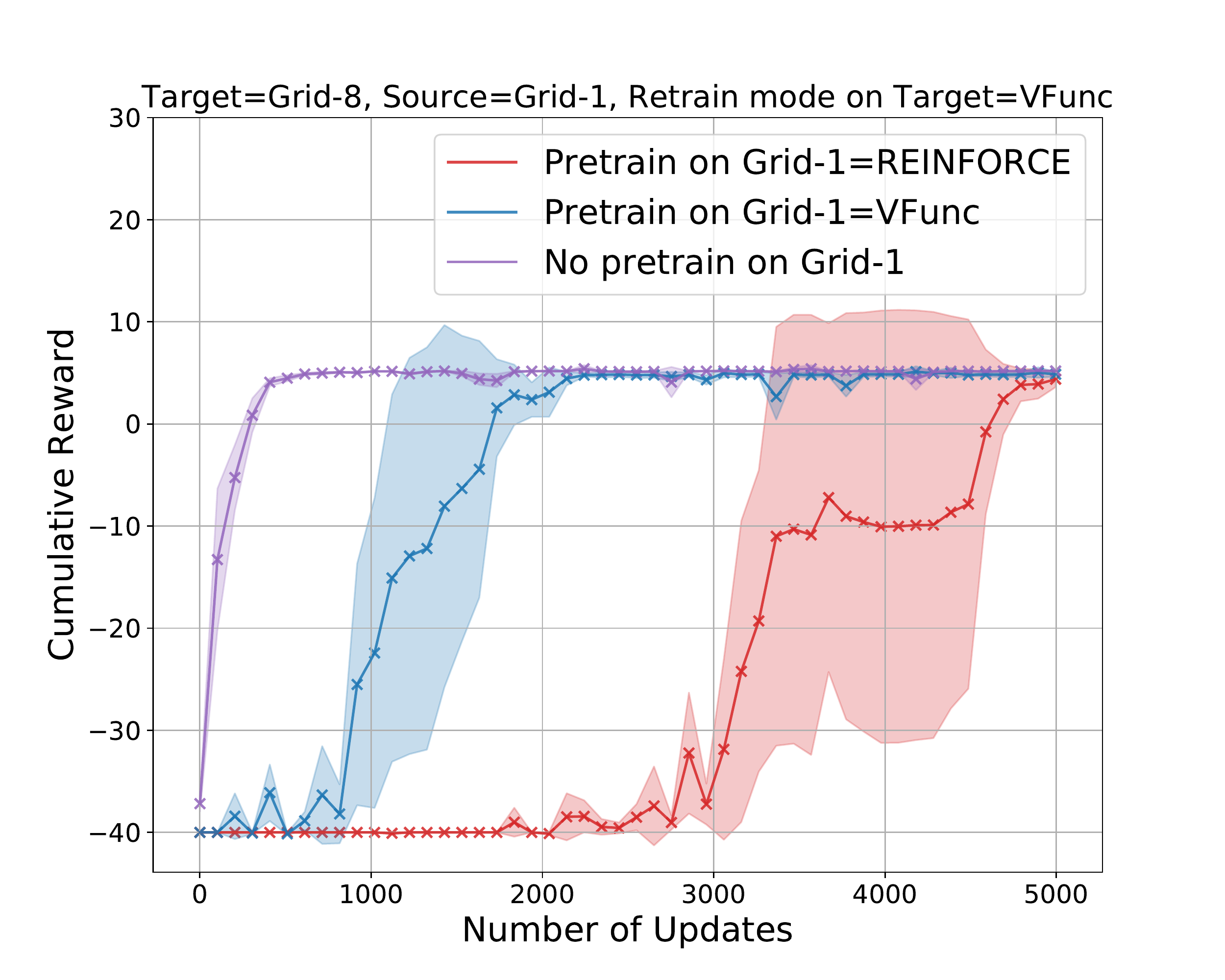}
            % \caption{Retrain with VFunc}
            \label{fig:vfunc_grid_3}
        \end{subfigure}
        \begin{subfigure}[b]{0.495\textwidth}
            \includegraphics[trim = {1.0cm 0 1.5cm 0}, clip, width=\textwidth]{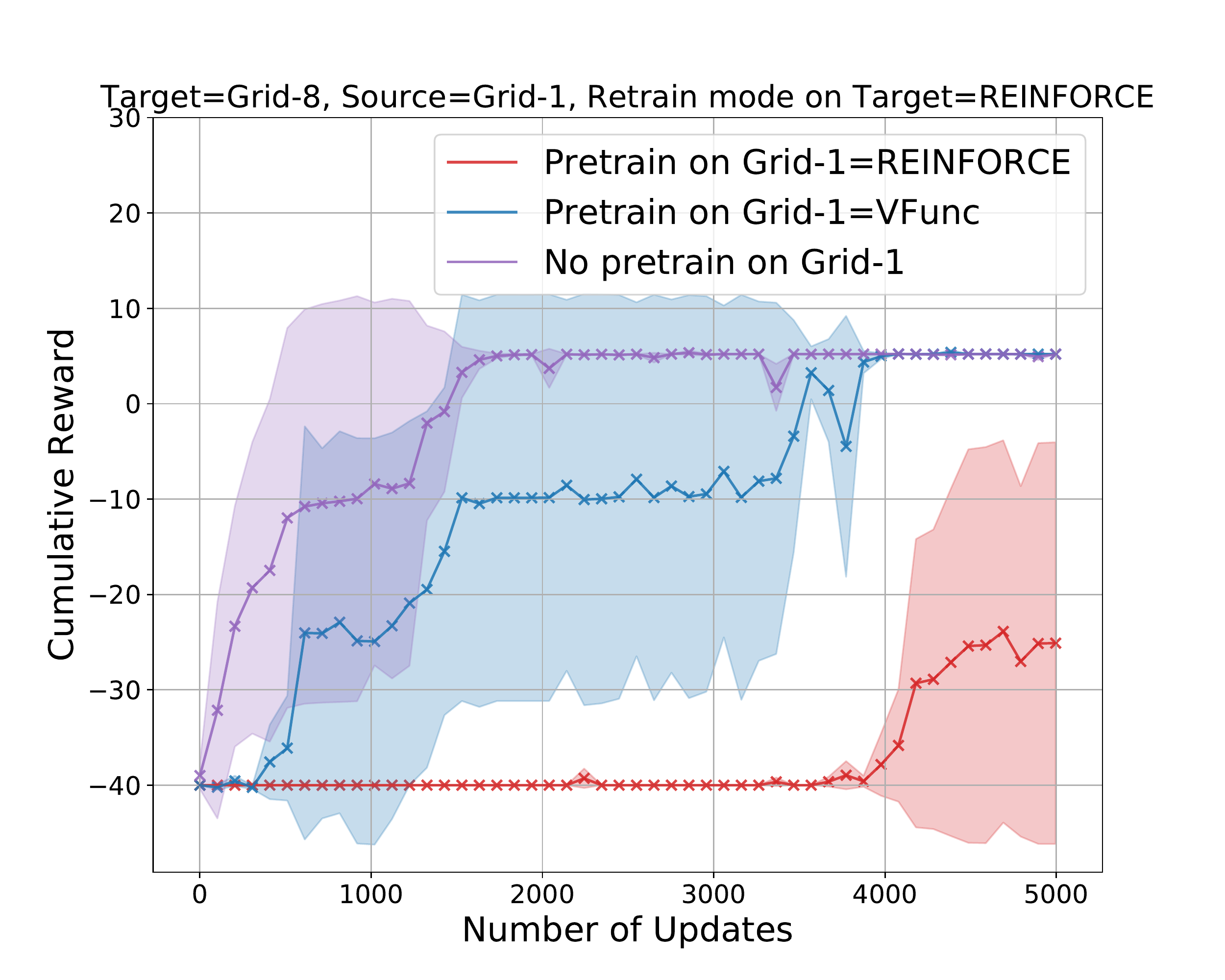}
            % \caption{Retrain with REINFORCE}
            \label{fig:r_grid_3}
        \end{subfigure}
        \vspace*{-1cm}
        \caption{Transfer on Grid-8}
         \label{transfer_grid_8}
    \end{subfigure}
    \caption{\textbf{Transfer Results on GridWorld:} Source env = Grid-1. Blue curve = Pretrained with VFunc, Purple curve = training from scratch, Red curve = Pretrained with REINFORCE. In each subfigure, left and right figures corresponds to retraining using VFunc and REINFORCE in the target environment, respectively. In almost all cases, pretraining with VFunc leads to faster learning of target policy.}
    \label{fig:transfer_gridworld}
\end{figure*}
%%%%%%%%%%%%%%%%%%%%%%%%%%%%%%%%%%%%%%%%%%%%%%%%%%
%%%%%%%%%%%%%%%%%%%%%%%%%%%%%%%%%%%%%%%%%%%%%%%%%%

%%%%%%%%%%%%%%%%%%%%%%%%%%%%%%%%%%%%%%%%%%%%%%%%%%
%%%%%%%%%%%%%%%%%%%%%%%%%%%%%%%%%%%%%%%%%%%%%%%%%%
% HEATMAPS
\begin{figure}[h]
    % ROW 1
    \centering
    \begin{subfigure}[h]{0.5\textwidth}
        \includegraphics[trim = {0.25cm 1.4cm 0.25cm 0.5cm}, clip, width=\textwidth]{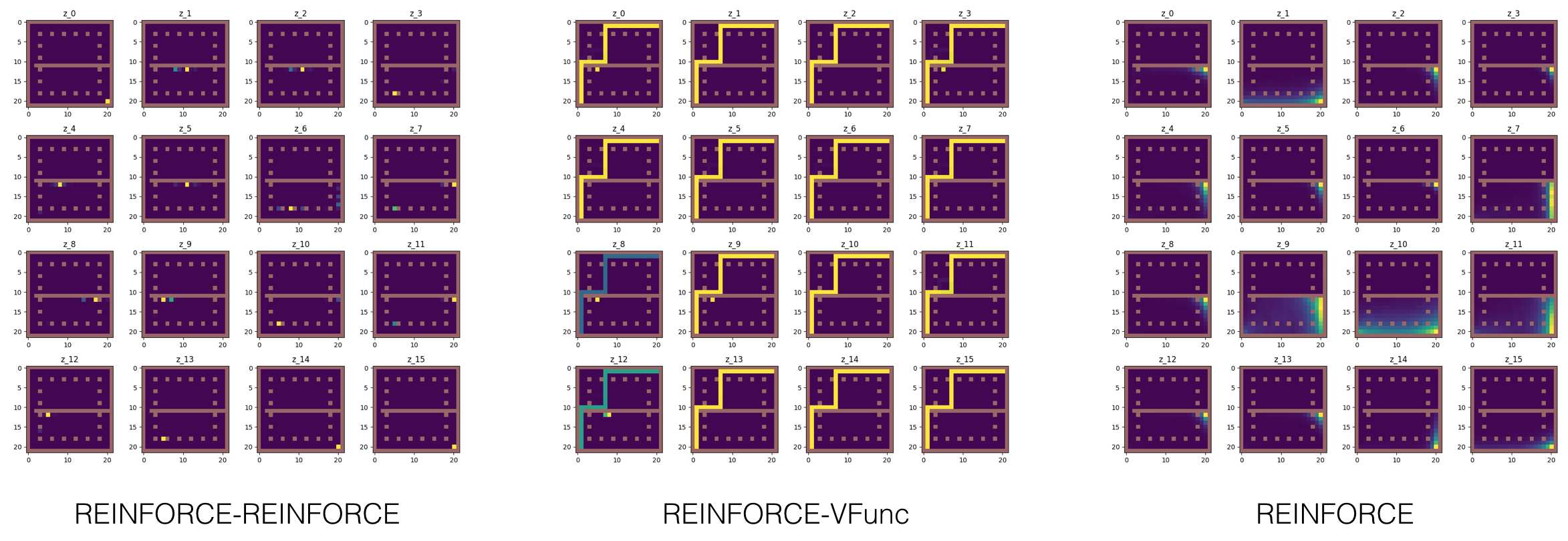}
        \label{fig:row1}
        \vspace*{-0.5cm}
        \caption{Sample heatmaps for Transfer with REINFORCE on Grid-2}
    \end{subfigure}
    
    \bigskip
    
    % ROW 2
    \centering
    \begin{subfigure}[h]{0.5\textwidth}
        \includegraphics[trim = {0.25cm 1.4cm 0.25cm 0.5cm}, clip, width=\textwidth]{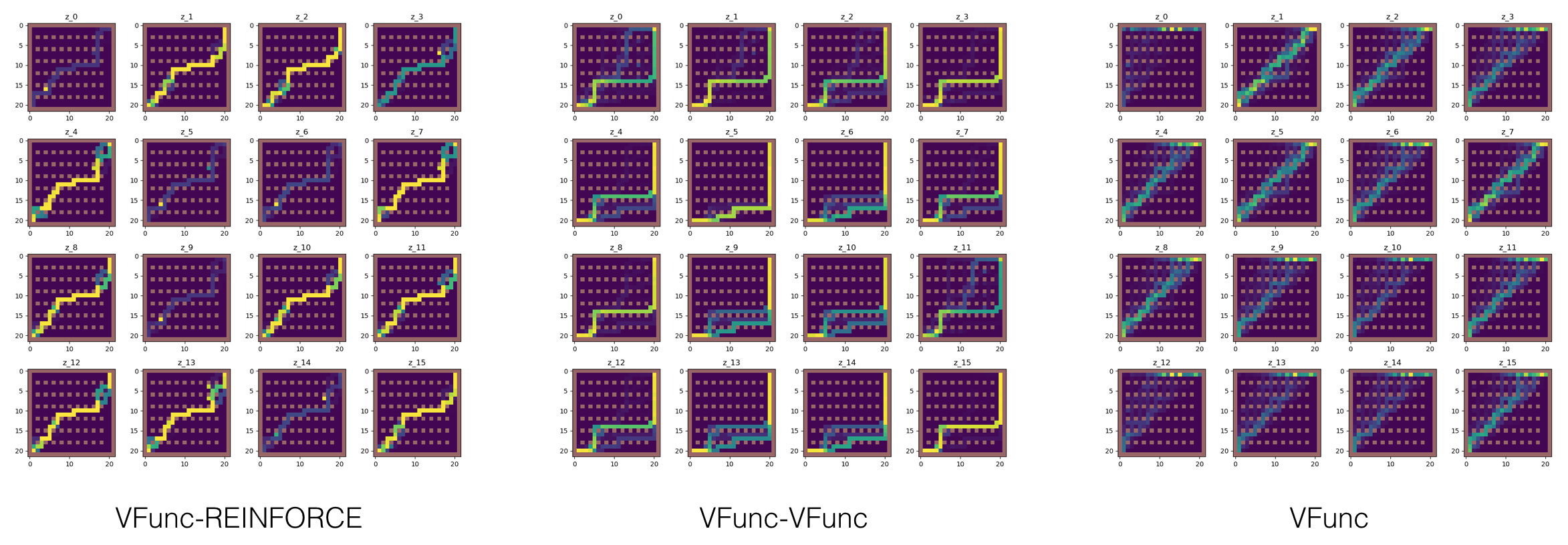}
        \label{fig:row2}
        \vspace*{-0.5cm}
        \caption{Sample heatmaps for Transfer with VFunc on Grid-3}
    \end{subfigure}
    
    \bigskip
    
    % ROW 3
    \centering
    \begin{subfigure}[h]{0.5\textwidth}
        \includegraphics[trim = {0.25cm 1.4cm 0.25cm 0.5cm}, clip, width=\textwidth]{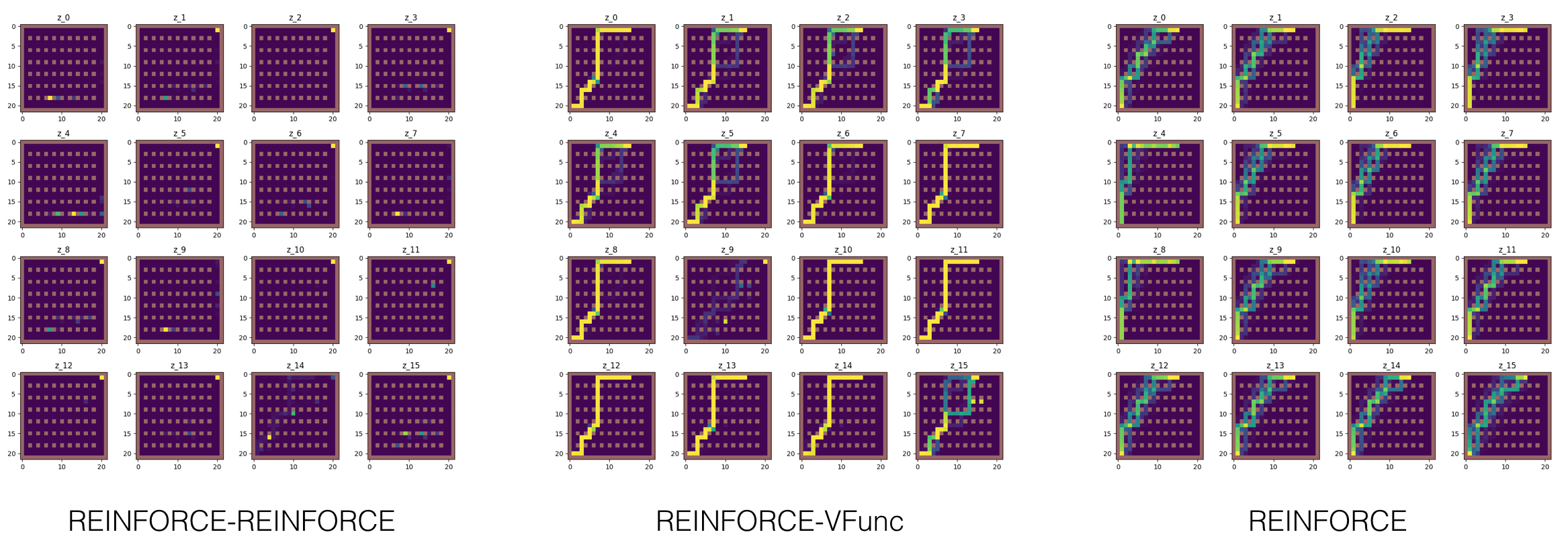}
        \label{fig:row3}
        \vspace*{-0.5cm}
        \caption{Sample heatmaps for Transfer with REINFORCE on Grid-7}
    \end{subfigure}

    \caption{\textbf{Heatmaps showing State Visitation Frequencies for 16 sampled policies for GridWorld}. The first and second column in each row corresponds to the case when Grid-1(source) is trained with REINFORCE and VFunc, respectively. The third column corresponds to training from scratch. Pre-training with VFunc results in policies which find successful paths between the start and goal states and at the same time are more diverse.}
    \label{heatMapsGridWorld}
\end{figure}
%%%%%%%%%%%%%%%%%%%%%%%%%%%%%%%%%%%%%%%%%%%%%%%%%%
%%%%%%%%%%%%%%%%%%%%%%%%%%%%%%%%%%%%%%%%%%%%%%%%%%

\subsection{Experiments}
\subsubsection{Comparison with REINFORCE on GridWorld}
We built upon the code base provided for VFunc by~\cite{bachman2018vfunc} and modified it to carry out transfer on different environments. In all experiments we took Grid-1 \ref{fig:grid-1} as the source environment where training is carried out using both REINFORCE~\cite{williams1992simple} and VFunc and the corresponding weights are stored. We then tested the performance on five different target environments(Grid-2, Grid-3, Grid-6, Grid-7 and Grid-8) with varying levels of difficulty and variation from the source environment. As can be seen from Figure~\ref{fig:gridworld}, we test on target environments where we add a horizontal wall in Grid-2, more vertical walls in Grid-3 and a combination of both horizontal and vertical walls in Grid-6. Grid-7 and Grid-8 are same as Grid-3 and Grid-6 respectively, but with the position of the goal state changed. We trained the source environment with both REINFORCE and VFunc (See Algorithm 1 and description from Figure~\ref{fig:formulation:arch}). This resulted in two sets of weights which were then used to initialize the network weights in the target environment. We then retrained the target environment with both REINFORCE and VFunc. For better comparison, we also trained the target environment from scratch (without initializing the network weights from the source environment) using both REINFORCE and VFunc.

\subsubsection{Comparison with A2C on MiniGrid}
We implemented Advantage Actor-Critic (called A2C henceforth) by~\cite{mnih2016asynchronous} as the policy-gradient algorithm and VFunc on top of it. The source environment was N2S4, i.e., a series of two rooms each of size four which was trained both with A2C and VFunc. We observed the performance while transferring the weights to N2S6 (two rooms of size six) and N3S4 (three rooms of size four) as the target environments and retraining using both A2C and VFunc. As before, we also trained the target environments from scratch without any weights initialized from the weights learned during training on the source environment. We experimented with both the \textbf{\textit{Static}} and \textbf{\textit{Dynamic}} settings.
%Please note that the environments here are designed to be dynamic, i.e., we get a different environment with same configuration for different episodes and across different parallel roll-outs.

%%%%%%%%%%%%%%%%%%%%%%%%%%%%%%%%%%%%%%%%%%%%%%%%%%
%%%%%%%%%%%%%%%%%%%%%%%%%%%%%%%%%%%%%%%%%%%%%%%%%%
\begin{figure*}[h]
    \centering
    % ROW 1
    \begin{subfigure}[h]{0.495\textwidth}
        \begin{subfigure}[b]{0.495\textwidth}
            \includegraphics[trim = {0.75cm 0 1.0cm 0}, clip, width=\textwidth]{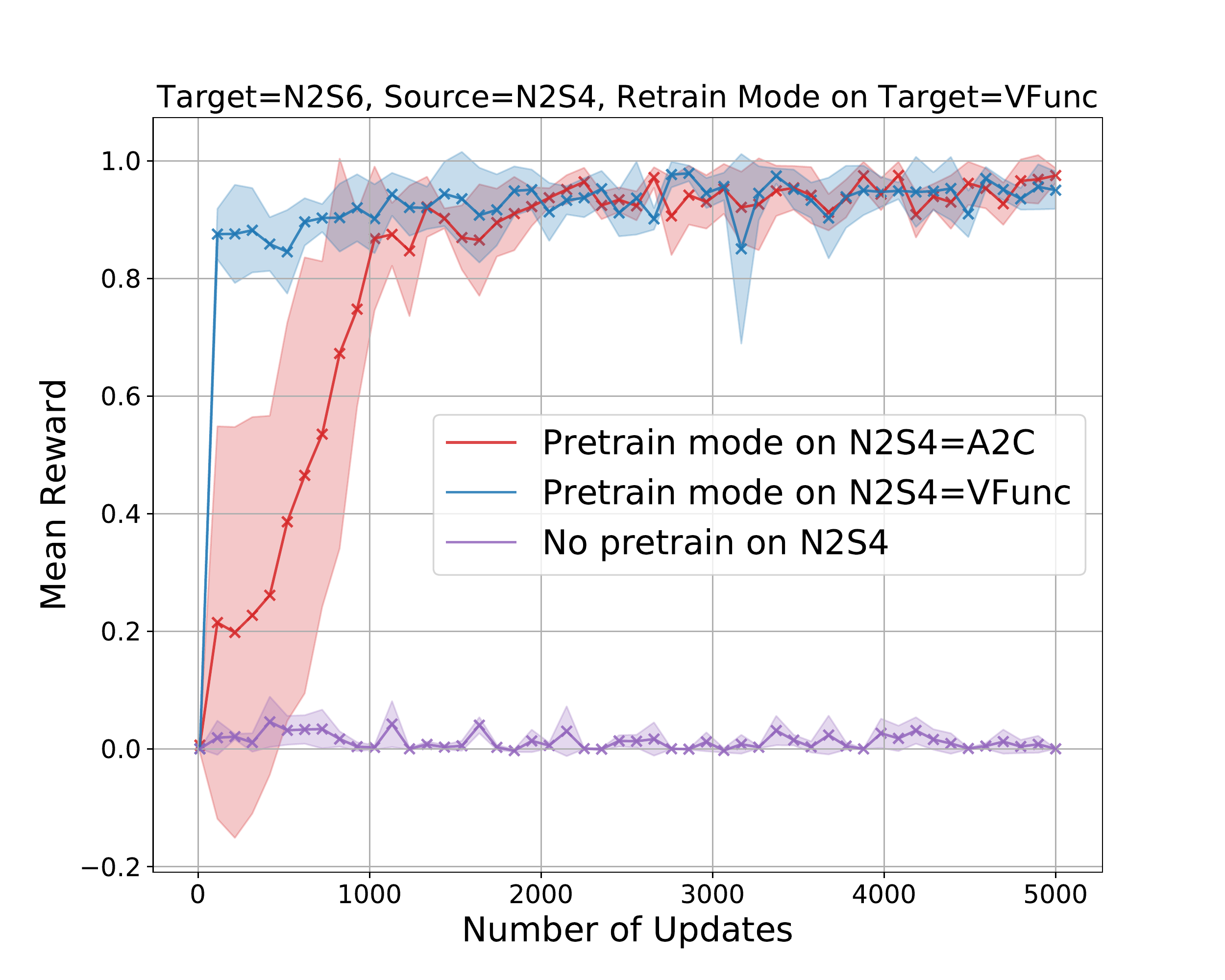}
            % \caption{Retrain with VFunc}
            \label{fig:vfunc_n2s6}
        \end{subfigure}
        \begin{subfigure}[b]{0.495\textwidth}
            \includegraphics[trim = {0.75cm 0 1.0cm 0}, clip, width=\textwidth]{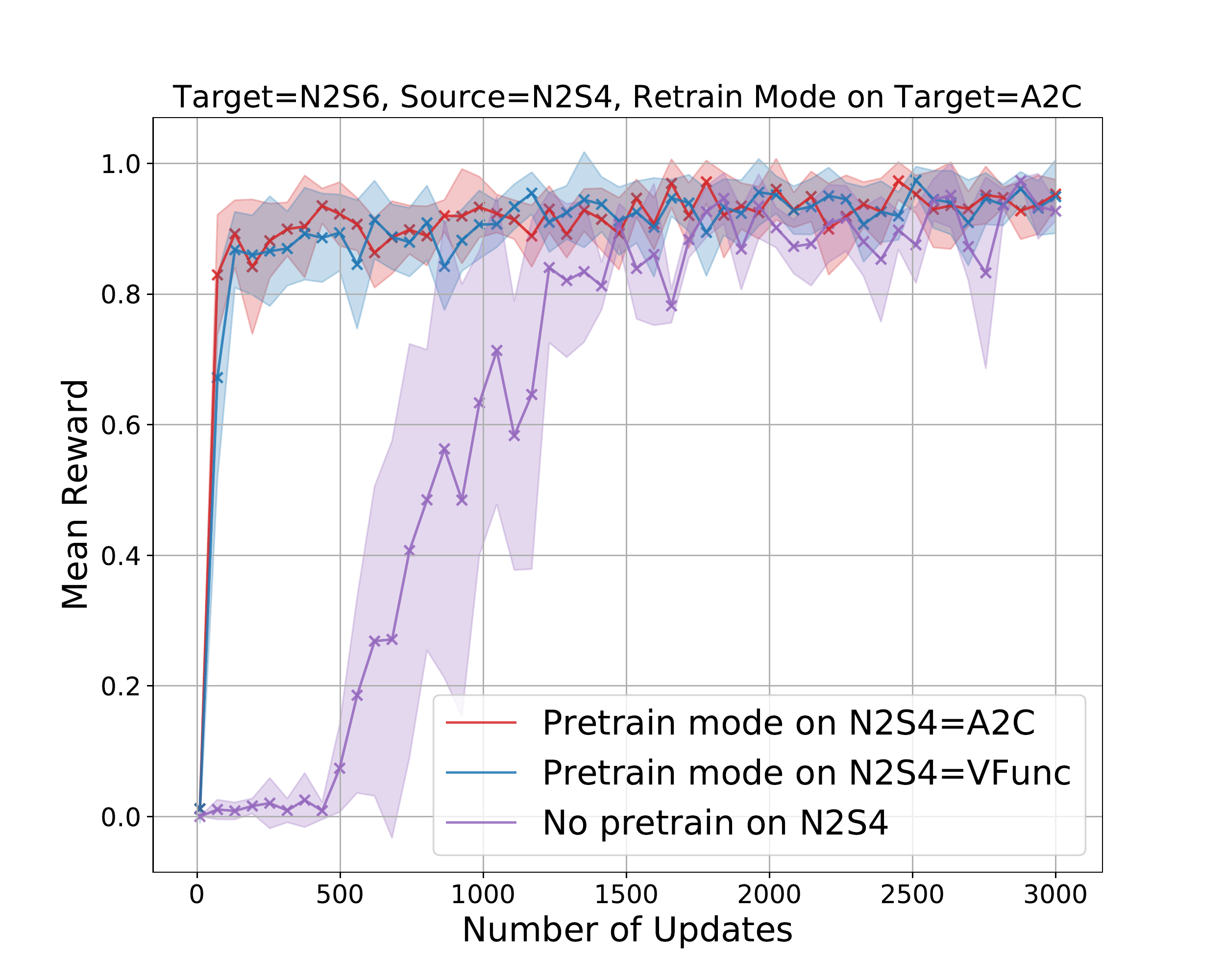}
            % \caption{Retrain with REINFORCE}
            \label{fig:a2c_n2s6}
        \end{subfigure}
        \vspace*{-1cm}
        \caption{Transfer on N2S6 (Dynamic)}
         \label{transfer_n2s6_d}
    \end{subfigure}
    \centering
    \begin{subfigure}[h]{0.495\textwidth}
        \begin{subfigure}[b]{0.495\textwidth}
            \includegraphics[trim = {0.75cm 0 1.0cm 0}, clip, width=\textwidth]{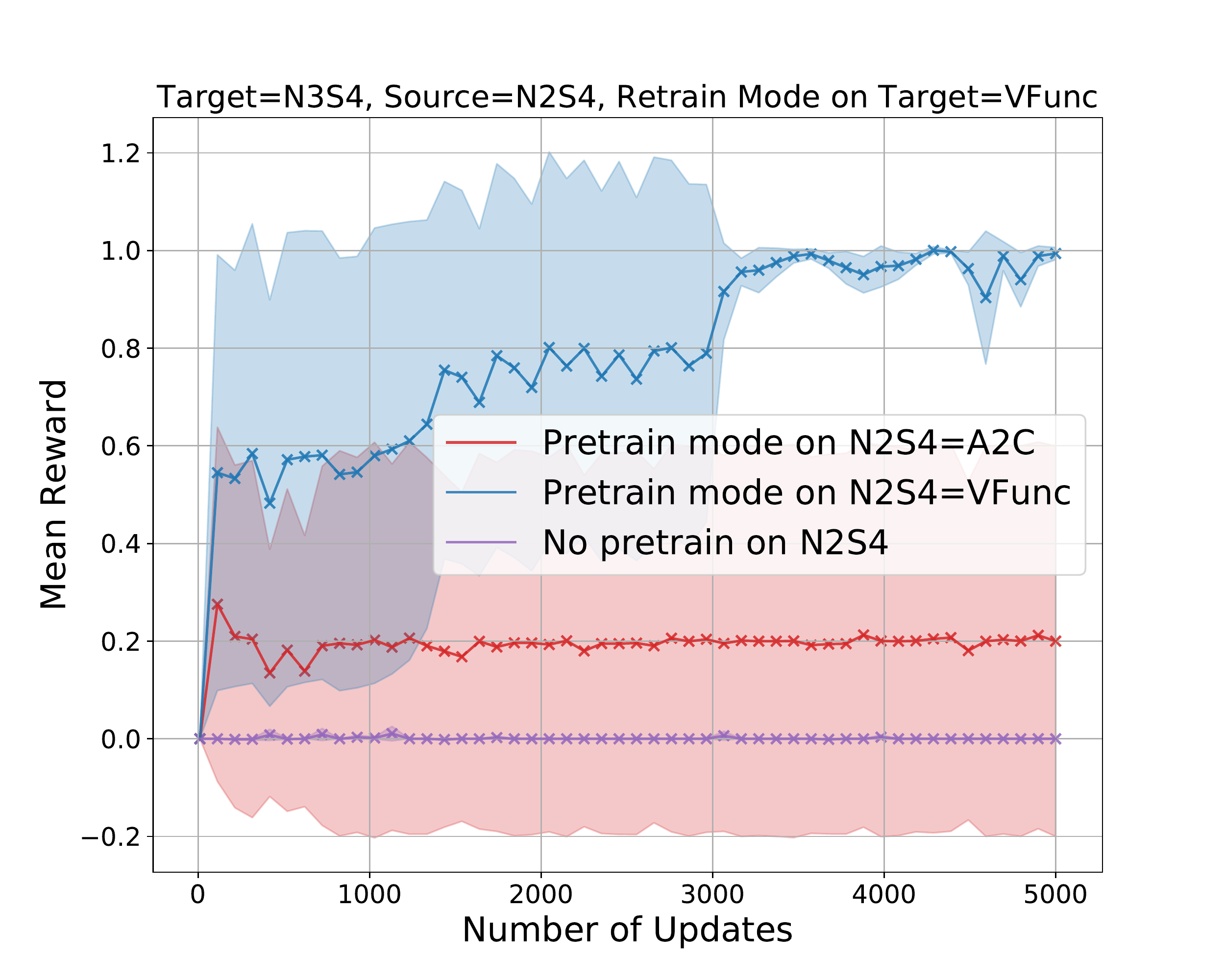}
            % \caption{Retrain with VFunc}
            \label{fig:vfunc_n3s4}
        \end{subfigure}
        \begin{subfigure}[b]{0.495\textwidth}
            \includegraphics[trim = {0.75cm 0 1.0cm 0}, clip, width=\textwidth]{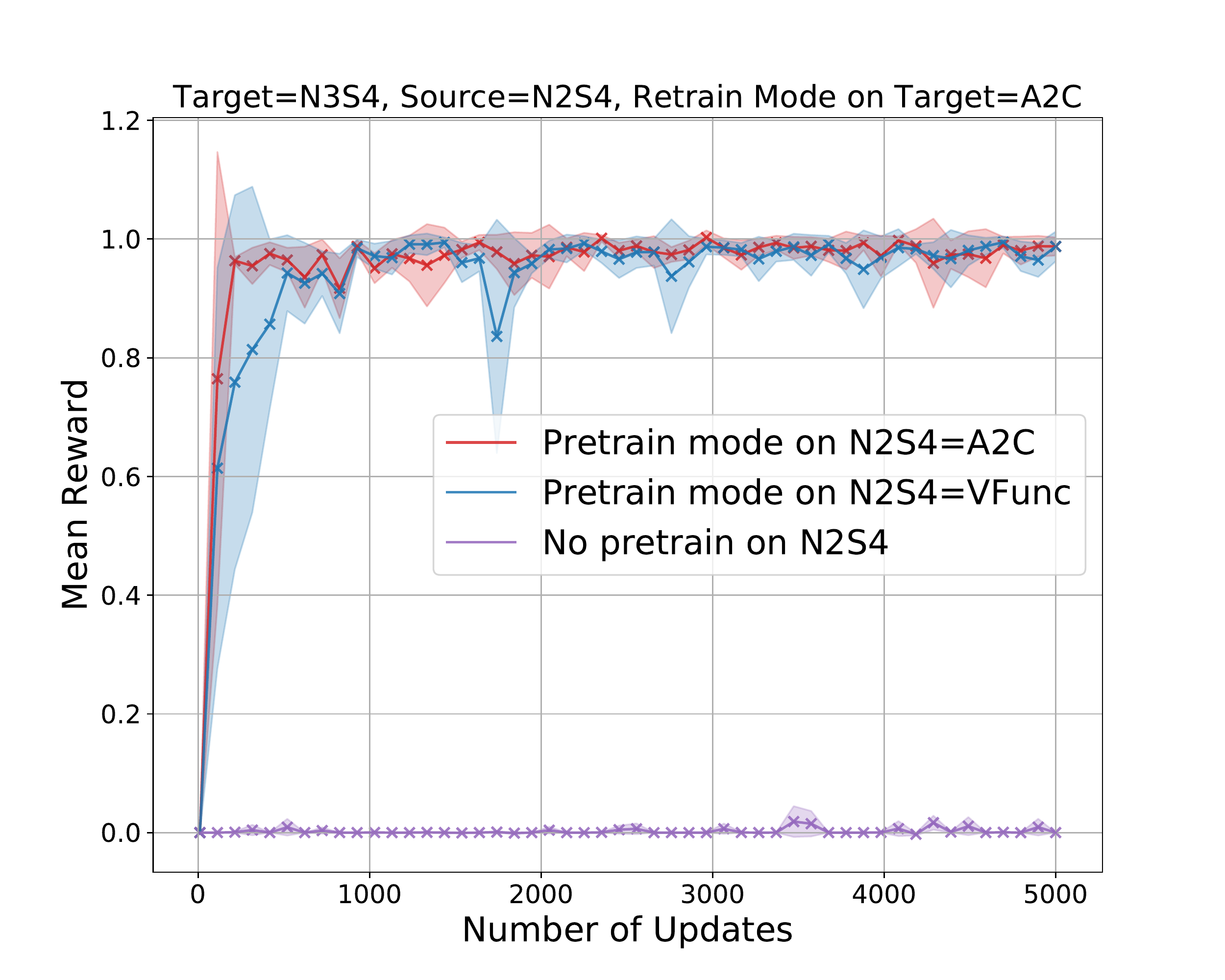}
            % \caption{Retrain with REINFORCE}
            \label{fig:a2c_n3s4}
        \end{subfigure}
        \vspace*{-1cm}
        \caption{Transfer on N3S4 (Dynamic)}
        \label{transfer_n3s4_d}
    \end{subfigure}
    % ROW 2
    \bigskip
    \centering
     \begin{subfigure}[h]{0.495\textwidth}
        \begin{subfigure}[b]{0.495\textwidth}
            \includegraphics[trim = {1.0cm 0 1.0cm 0}, clip, width=\textwidth]{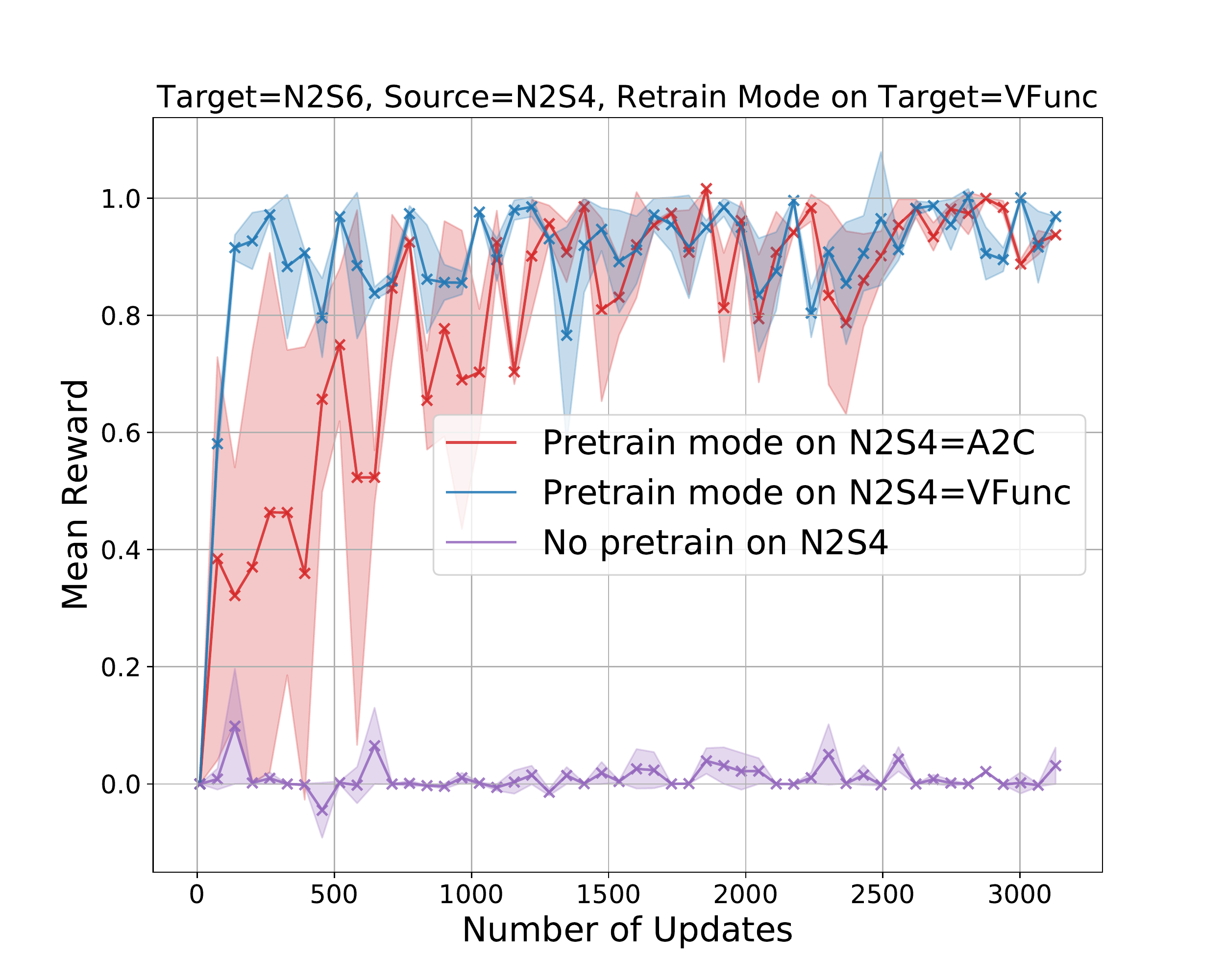}
            % \caption{Retrain with VFunc}
            \label{fig:vfunc_n2s6_static}
        \end{subfigure}
        \begin{subfigure}[b]{0.495\textwidth}
            \includegraphics[trim = {1.0cm 0 1.0cm 0}, clip, width=\textwidth]{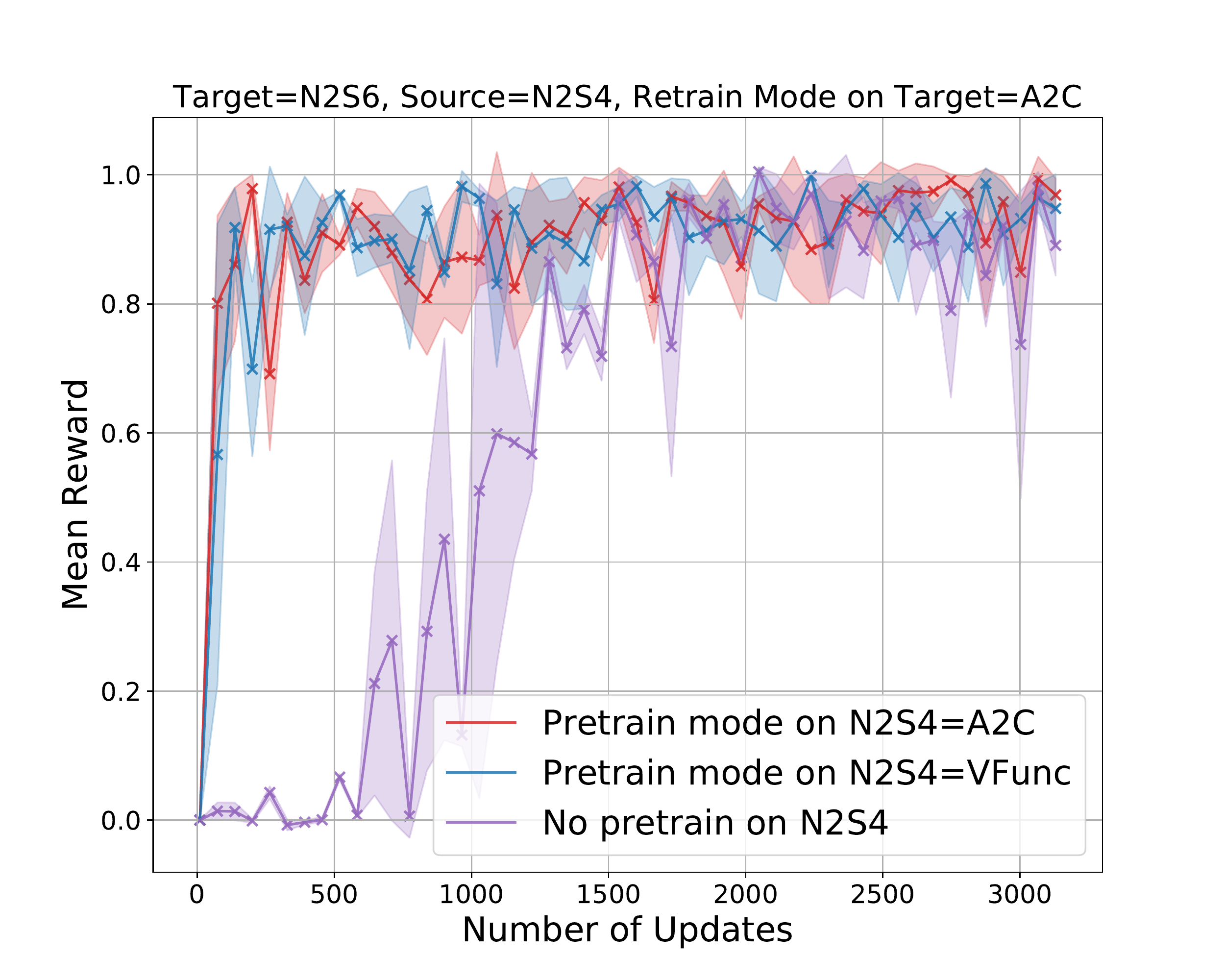}
            % \caption{Retrain with REINFORCE}
            \label{fig:a2c_n2s6_static}
        \end{subfigure}
        \vspace*{-1cm}
        \caption{Transfer on N2S6 (Static)}
         \label{transfer_n2s6_s}
    \end{subfigure}
    \centering
    \begin{subfigure}[h]{0.495\textwidth}
        \begin{subfigure}[b]{0.495\textwidth}
            \includegraphics[trim = {1.0cm 0 1.0cm 0}, clip, width=\textwidth]{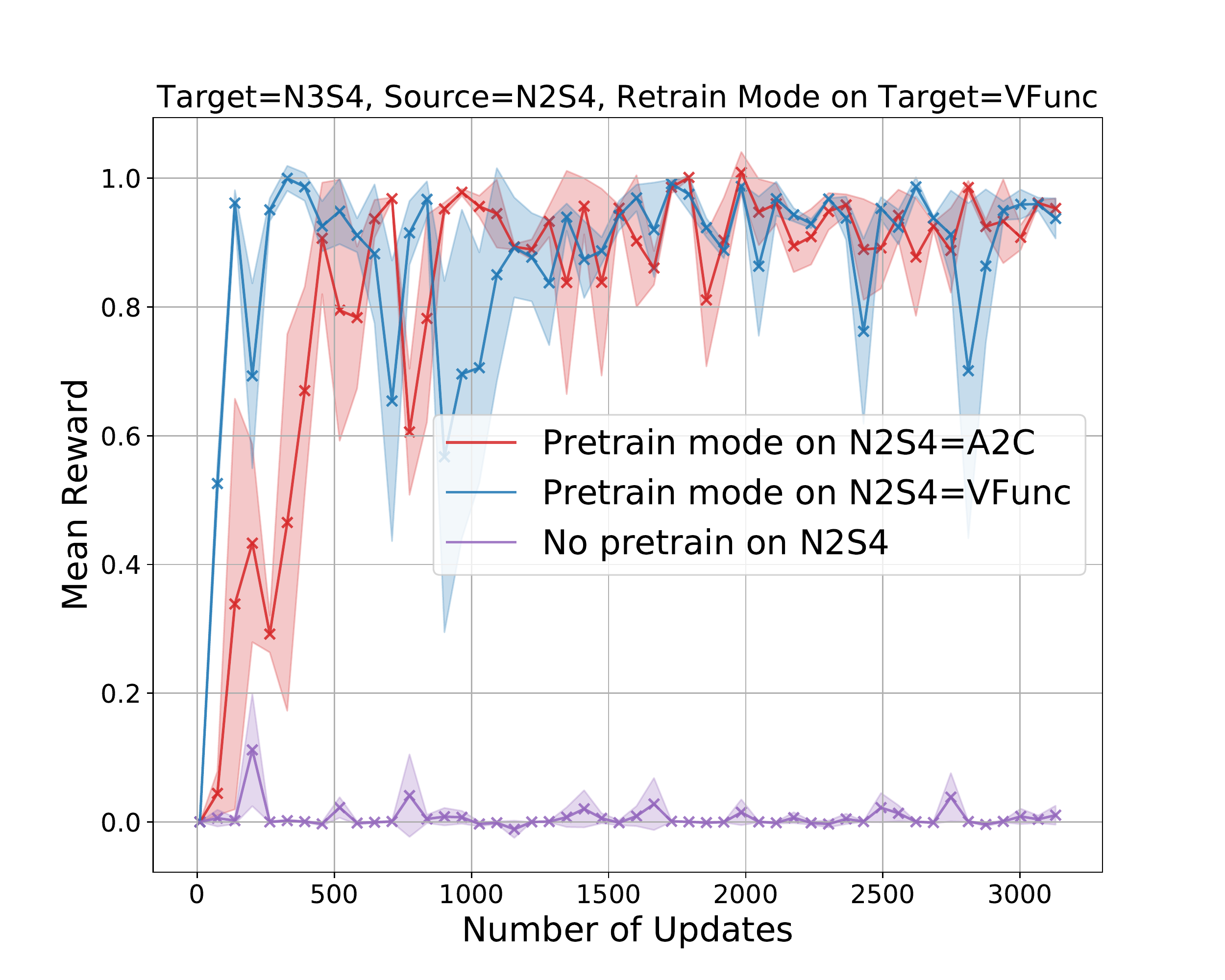}
            % \caption{Retrain with VFunc}
            \label{fig:vfunc_n3s4_static}
        \end{subfigure}
        \begin{subfigure}[b]{0.495\textwidth}
            \includegraphics[trim = {1.0cm 0 1.0cm 0}, clip, width=\textwidth]{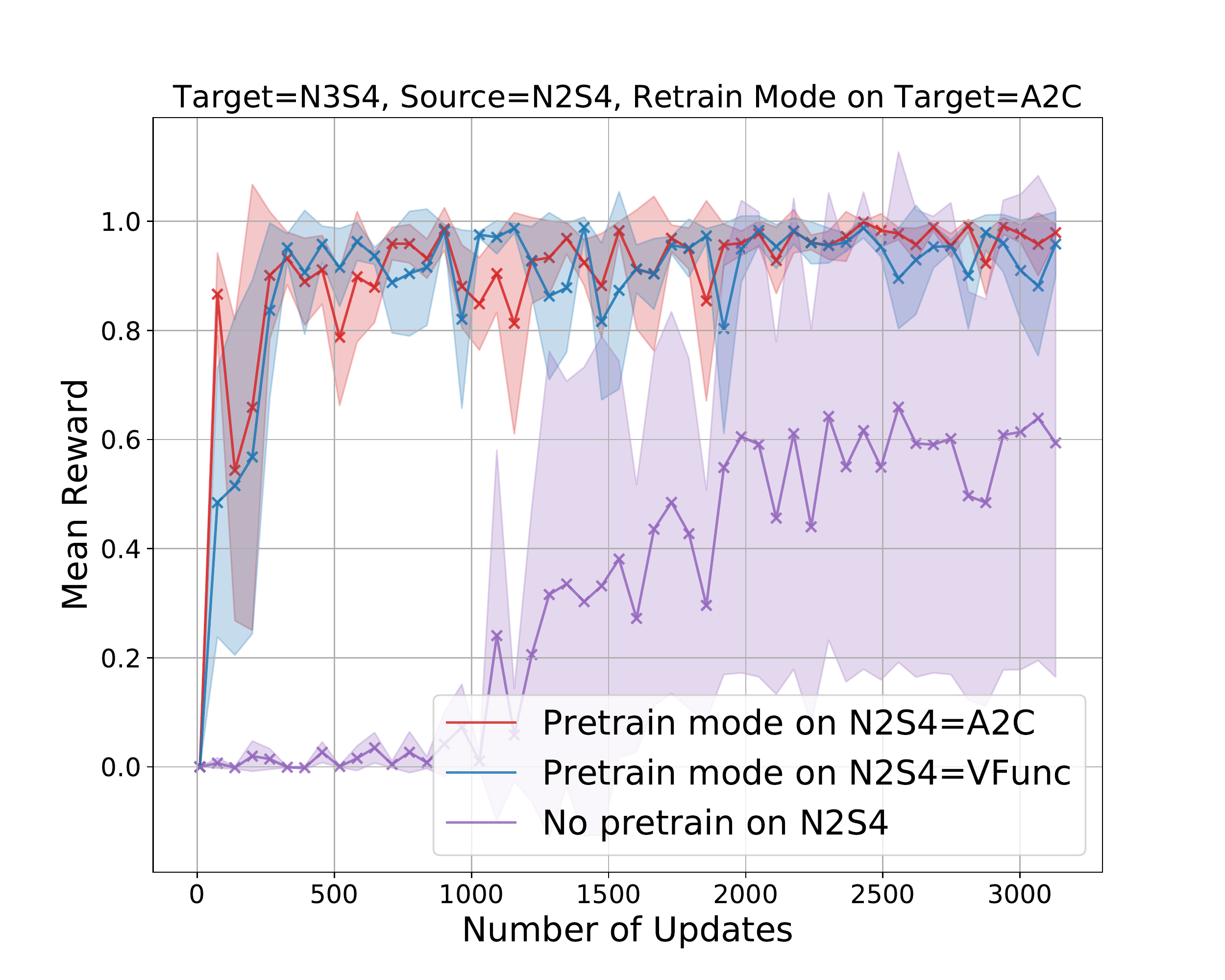}
            % \caption{Retrain with REINFORCE}
            \label{fig:a2c_n3s4_static}
        \end{subfigure}
        \vspace*{-1cm}
        \caption{Transfer on N3S4 (Static)}
        \label{transfer_n3s4_s}
    \end{subfigure}
    \caption{\textbf{Transfer Results on MiniGrid:} Source env = N2S4. Blue curve = Pretrained with VFunc, Purple curve = training from scratch, Red curve = Pretrained with A2C. In each subfigure, left and right figures corresponds to retraining using VFunc and A2C in the target environment, respectively. In both static and dynamic settings and both target environments, pretraining with VFunc is useful for learning target policy.}
    \label{fig:transfer_minigrid}
\end{figure*}
%%%%%%%%%%%%%%%%%%%%%%%%%%%%%%%%%%%%%%%%%%%%%%%%%%
%%%%%%%%%%%%%%%%%%%%%%%%%%%%%%%%%%%%%%%%%%%%%%%%%%

\section{Results and Discussions}
\subsection{Results on GridWorld}

The performance was measured by plotting the cumulative reward averaged across episodes and parallel processes with respect to the number of updates. We perform runs with 3 or more random seeds. Figures \ref{fig:transfer_gridworld} shows the results for transfer to Grid-2, Grid-3, Grid-6, Grid-7 and Grid-8 after training on Grid-1 as source; as well as training on these target environments from scratch. In the plots, the blue curve corresponds to the case when the weights of the source environment trained on VFunc are used to do transfer on the target environment. The red curve corresponds to the same case except that the weights of the source environment come from REINFORCE. The purple curves correspond to training from scratch on the target environment. Shaded regions around each curve reflect the variability in runs carried out across multiple seeds. In terms of difficulty, Grid-2 should be the most difficult to do transfer learning as it differs the most from the source environment (Grid-1). It is also difficult to learn from scratch as there is only one unblocked path on the far left which leads from the start to goal state. As we can see from Figure~\ref{transfer_grid_2}, transferring with VFunc trained weights is able to learn fast. Training from scratch on Grid-2 is the worst. In fact, it is not able to learn anything useful. On Grid-3 and Grid-6, Figures \ref{transfer_grid_3} and \ref{transfer_grid_6}, VFunc performs the best. Grid-7 is comparatively easy to solve due to its high resemblance to the source environment and VFunc emerges as a clear winner here (Figure~\ref{transfer_grid_7}) as it is able to transfer the knowledge present in the distribution of policies learned in the source environment quite readily to the target environment. Grid-8 being of medium difficulty and the goal state changed makes VFunc not the best choice to do transfer learning. In fact, training from scratch is better here (Figure~\ref{transfer_grid_8}). 

In Figure \ref{heatMapsGridWorld}, we also show sample heatmaps for the state visitation frequency corresponding to roll-outs pertaining to 16 different samples values of the latent variable, $z$ for Grid-2, Grid-3 and Grid-7. We see that transfer with VFunc pretrained model leads to finding successful paths between the start and goal states in all cases. In some cases, pretraining with REINFORCE or training from scratch on target environment is not able to discover any path between the start and goal states. We also see that the target policies learned by VFunc are quite diverse (as indicated by occasional blueness in the trajectory heatmaps) suggesting its usefulness for exploration in the target environment.

\subsection{Results on MiniGrid}
Figures \ref{transfer_n2s6_d} and \ref{transfer_n3s4_d} shows the results of transfer experiments carried out on N2S6 and N3S4 using N2S4 as the source environment in the \textbf{\textit{Dynamic}} setting. Figures \ref{transfer_n2s6_s} and \ref{transfer_n3s4_s} represent the same for \textbf{\textit{Static}} setting. The description of the legends is the same as above. All experiments correspond to runs with 5 different random seeds. As can be seen from the plots, training on N2S4 with VFunc and transferring using VFunc converges in less number of updates for both the settings and for both environments: N2S6 and N3S4. When we retrain with A2C on the target environment, the performance of VFunc is at par with A2C trained weights on N2S4. Training from scratch in the target environments is useless here. We believe that better hyperparameter tuning will help VFunc outperform A2C pretraining for N3S4. Besides, transfer to more complex environments will showcase the benefits of VFunc in a more appealing way.

\section{Conclusions and Future Work}
We observe that learning the target policy when pre-training with VFunc leads to faster convergence as compared to pre-training with other policy-gradient techniques (like REINFORCE and A2C) in the source task as well as training from scratch on the target task. The difference is more apparent as the difficulty level of the environment increases in terms of its variation from the source environment and environment dynamics (partial observability, degree of stochasticity, etc.). The explanation for improved transfer performance of VFunc is that during training with VFunc on the source environment, the model learns a distribution of good policies to explore in the target environment. This leads to faster exploration and hence, faster convergence. In addition, the learned policies are more diverse.  In future, we wish to explore the Safe AI perspective \cite{jain2018safe} wherein the idea is that since VFunc learns a distribution over policies rather than a single optimal policy, it will be able to learn few policies which avoid dangerous states (states with a very high penalty). We plan to experiment with AI Safe GridWorlds in \cite{leike2017ai} for this task.

\bibliography{icml2019}
\bibliographystyle{icml2019}

\end{document}